%% file: eccv_latex.tex
\documentclass[runningheads]{llncs}

\usepackage{eccv}

\usepackage{graphicx}
\usepackage{booktabs}
\usepackage{booktabs}
\usepackage{multirow}

\usepackage{amsmath}
\usepackage{bbold} 

\usepackage{algorithmic} 
\usepackage{algorithm}

\DeclareMathAlphabet\mathbfcal{OMS}{cmsy}{b}{n}
\makeatletter
\def\@fnsymbol#1{\ensuremath{\ifcase#1\or \dagger\or *\or \ddagger\or
   \mathsection\or \mathparagraph\or \|\or **\or \dagger\dagger
   \or \ddagger\ddagger \else\@ctrerr\fi}}

\usepackage[accsupp]{axessibility}  

\usepackage[pagebackref,breaklinks,colorlinks,citecolor=eccvblue]{hyperref}

\usepackage{orcidlink}

\begin{document}










\title{Exploiting Semantic Reconstruction to Mitigate Hallucinations in Vision-Language Models}

\titlerunning{ESREAL}


\author{Minchan Kim$^1$\thanks{Equal contribution.}\index{Kim, Minchan} \and
Minyeong Kim$^1$\textsuperscript{$\dagger$} \and
Junik Bae$^1$\textsuperscript{$\dagger$} \and
Suhwan Choi$^1$ \and \\
Sungkyung Kim$^1$ \and
Buru Chang$^2$\thanks{Corresponding author.}}
\authorrunning{M. Kim and M. Kim et al.}
%
\institute{Seoul National University \and Sogang University\\
\email{\{kjkk0502,kmy17518,heatz123,milkclouds,sk0428\}@snu.ac.kr\\
buru@sogang.ac.kr}
}
\maketitle
\input{Sections/0_abstract}
\input{Sections/1_introduction}
\input{Sections/2_related_work}
\input{Sections/3_method}
\input{Sections/4_experiments}
\input{Sections/5_analysis}
\input{Sections/6_conclusion}

\input{Sections/aknowledgements}

\bibliographystyle{splncs04}
\bibliography{custom}
\input{Sections/A_appendix}

\end{document}

%% file: Sections/0_abstract.tex
\begin{abstract}\label{sec:0_abstract}

Hallucinations in vision-language models pose a significant challenge to their reliability, particularly in the generation of long captions.
Current methods fall short of accurately identifying and mitigating these hallucinations.
To address this issue, we introduce ESREAL, a novel unsupervised learning framework designed to suppress the generation of hallucinations through accurate localization and penalization of hallucinated tokens.
Initially, ESREAL creates a reconstructed image based on the generated caption and aligns its corresponding regions with those of the original image.
This semantic reconstruction aids in identifying both the presence and type of token-level hallucinations within the generated caption.
Subsequently, ESREAL computes token-level hallucination scores by assessing the semantic similarity of aligned regions based on the type of hallucination.
Finally, ESREAL employs a proximal policy optimization algorithm, where it selectively penalizes hallucinated tokens according to their token-level hallucination scores.
Our framework notably reduces hallucinations in LLaVA, InstructBLIP, and mPLUG-Owl2 by 32.81\%, 27.08\%, and 7.46\% on the CHAIR metric.
This improvement is achieved solely through signals derived from the image itself, without the need for any image-text pairs.

\end{abstract}

%% file: Sections/1_introduction.tex
\section{Introduction}\label{sec:1_introduction}

The evolution of Vision-Language Models (VLMs)~\cite{li2022blip,zhu2023minigpt} represents a substantial leap forward.
By integrating Large Language Models (LLMs)~\cite{vicuna2023} as core components, modern VLMs~\cite{liu2023visual, dai2023instructblip} are capable of handling diverse multimodal tasks such as Visual Question Answering (VQA) and image captioning.
Despite these advancements, VLMs still suffer from the notorious \textit{hallucination} problem.
Hallucination refers to the phenomenon where the VLM generates captions that are either incorrect or inconsistent with the given image and prompt.
Hallucinations greatly undermine the reliability of VLMs, particularly in real-world implementations where accuracy is paramount~\cite{huang2023voxposer, hu2023avis}.

\input{Figures/1_motivation}

Recent works have proposed training methods for VLMs to mitigate the hallucination problem.
A prominent strategy involves the use of additional datasets designed to combat hallucinations, which are created by GPT models~\cite{liu2023mitigating} or annotated by humans~\cite{sun2023aligning,gunjal2023detecting}.
Current data-driven approaches detect hallucinations at the sentence or paragraph levels, overlooking detailed aspects such as specific locations and types of hallucinations.
Despite the potential benefits of these fine-grained details for understanding and mitigating hallucinations, annotating data at this level is expensive.

Therefore, we introduce ESREAL\footnote{The abbreviation of \underline{E}xploiting \underline{S}emantic \underline{RE}construction to mitigate h\underline{AL}lucinations}, a fully unsupervised learning framework designed to mitigate hallucinations in VLMs, especially in generating long captions.
ESREAL exploits the token-level location and type of hallucinations without the need for annotated data.
To make this feasible, ESREAL incorporates a reference-free hallucination detection pipeline utilizing \textit{semantic reconstruction}.
Our hallucination detection pipeline begins by semantically reconstructing the original image from the generated caption with a text-to-image model.
As shown in Figure~\ref{fig:1_motivation}, semantic misalignment between the two images emerges due to the precise reflection of hallucinated tokens in the reconstructed image.
Hence, our detection pipeline identifies regions of semantic misalignment and calculates scores for the corresponding hallucinated tokens by comparing these regions according to the hallucination type.

Subsequently, ESREAL leverages the scores produced by the hallucination detection pipeline as token-level penalties with a fine-grained Proximal Policy Optimization (PPO)~\cite{schulman2017proximal} approach.
By selectively penalizing the hallucinated tokens, ESREAL can successfully suppress the generation of hallucinatory content.
We conduct experiments to evaluate ESREAL on three open-source VLMs. 
Our proposed framework achieves a 32.81\%, 27.08\%, 7.46\% improvement in the CHAIR metric over LLaVA~\cite{liu2023llava}, InstructBLIP~\cite{dai2023instructblip}, and mPLUG-Owl2~\cite{ye2023mplug}, respectively.
Additionally, ESREAL consistently enhances the performance of VLMs on more comprehensive model-based evaluation methods such as FaithScore~\cite{jing2023faithscore} and GPT-4V-aided evaluation~\cite{zhang2023gpt4vision}.

\textbf{Contributions}. 
Our contributions are threefold.
First, we propose ESREAL, a fully unsupervised hallucination mitigation framework. Our approach is scalable, eliminating the need for annotated data during training.
Second, we craft a hallucination detection pipeline, which facilitates token-level identification of hallucinations in generated captions in a reference-free manner via semantic reconstruction.
Finally, we show that ESREAL can be applied across a variety of VLMs to effectively mitigate hallucinations.

%% file: Figures/1_motivation.tex
\begin{figure}[!t]
\centering
\includegraphics[width=0.95\textwidth]{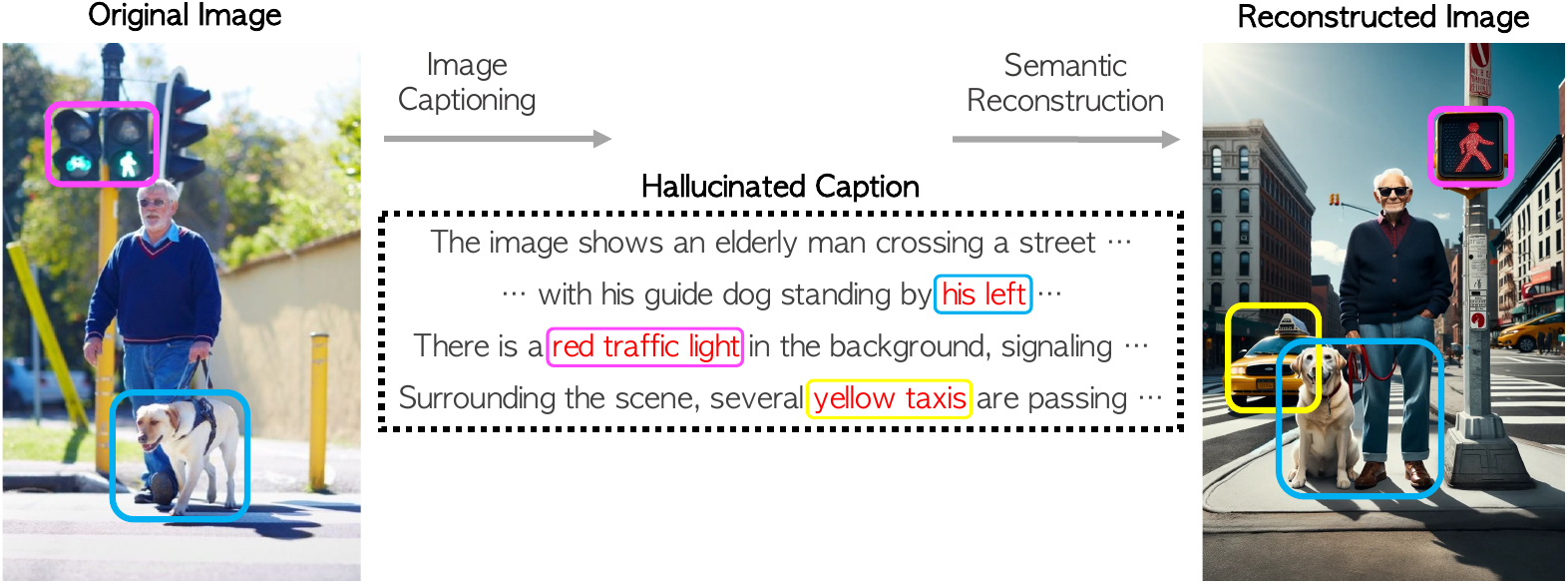}
\caption{Motivation of our study. Hallucinated tokens within the caption lead to semantic misalignment between the original and reconstructed images. By comparing the disparities among corresponding regions in the images, we can effectively identify and localize the hallucinated tokens within the caption.
}
\label{fig:1_motivation}
\end{figure}

%% file: Sections/2_related_work.tex
\section{Related Work}\label{sec:2_related_work}

\subsection{Vision-Language Models}

The advent of LLMs~\cite{brown2020language,touvron2023llama} propelled the development of VLMs, integrating LLMs within a multimodal framework. 
These models commonly incorporate a visual encoder, a modality connection module, and a LLM. 
The visual encoder produces an abstraction of the visual input data, which is then bridged with the linguistic domain by the modality connection module. Subsequently, the LLM performs causal sequence generation conditioned on the visual input and prompt.
Modern VLMs are characterized by the architecture of the modality connection module.  
LLaVA~\cite{liu2023visual} adopts a simple linear layer while InstructBLIP~\cite{dai2023instructblip} inherits the Querying Transformer of BLIP-2~\cite{li2023blip2} to align the visual inputs with the LLM.
mPLUG-Owl2~\cite{ye2023mplug} introduces a modality-adaptive module that preserves modality-specific features while integrating visual and language features.

\subsection{Hallucination Mitigation in VLMs}

Recent studies have attempted to mitigate hallucinations with a data-driven approach.
One major trend is generating hallucination data with GPT-4~\cite{achiam2023gpt} and fine-tuning VLMs to distinguish between accurate and hallucinated responses.
GAVIE~\cite{liu2023mitigating} introduces the LRV dataset, incorporating negative prompts designed to elicit ``no'' responses from VLMs when hallucinations are detected. 
HACL~\cite{jiang2024hallucination} applies contrastive learning techniques with augmented data to combat hallucinations.
Utilizing human-annotated data for training reward models and applying reinforcement learning techniques to VLMs represents another emerging trend.
MMHal-Bench~\cite{sun2023aligning} ranks VLM responses based on the level of hallucination, adapting the reinforcement learning from human feedback~\cite{ouyang2022training} to the vision-language domain. 
MHalDetect employs direct preference optimization~\cite{rafailov2023direct} with a sentence-level hallucination detection model as its reward mechanism.

In this paper, our aim is to conduct a thorough investigation into the token-level location and types of hallucinations. 
The substantial costs associated with annotating such detailed information present a notable challenge for previous data-driven methodologies. 
Therefore, we introduce a novel unsupervised framework, ESREAL. Our framework directs models to mitigate hallucinations by imposing specific penalties on identified hallucinated tokens, without requiring additional data.

\subsection{Hallucination Evaluation in VLMs}

Several studies aim to evaluate the performance of VLMs in terms of hallucination. Some approaches assess a VLM's capability to recognize hallucinations through a separate VQA benchmark. Techniques such as POPE~\cite{li2023evaluating} and NOPE~\cite{lovenia2023negative} are designed to evaluate a model's proficiency in discriminating non-existent objects.
Other approaches aim to quantify hallucinations within a model's response. 
CHAIR~\cite{rohrbach2018object} measures the frequency of non-existent objects in generated responses by comparing objects extracted from them with ground-truth objects. 
FaithScore~\cite{jing2023faithscore} breaks down the generated response into atomic facts via powerful LLMs\cite{achiam2023gpt} and applies a visual entailment model to judge whether these atomic facts are supported by the visual content.

%% file: Sections/3_method.tex
\section{Method}\label{sec:3_method}

\input{Figures/2_overview}

\subsection{Overview of ESREAL}\label{subsec:3_1_overview}

In this section, we introduce ESREAL, an unsupervised learning framework designed to mitigate hallucinations in long captions.
ESREAL stems from the idea that providing fine-grained negative feedback on hallucinated tokens can deter their generation without impairing the generative abilities of VLMs. 
To this end, we develop a reference-free hallucination detection pipeline to pinpoint hallucinated tokens in the generated captions. 
This mechanism is then employed as the reward model in a fine-grained PPO algorithm~\cite{wu2023finegrained}. 
Figure~\ref{fig:2_overview} provides an overview of our framework.

\subsection{Reference-free Hallucination Detection Pipeline}\label{subsec:3_2_detection_pipeline}

The hallucination detection pipeline is responsible for locating hallucinated tokens and producing appropriate scores for distinct types of hallucinations. 
Following prior works~\cite{wang2023llm, wang2024amber}, we classify hallucinations into three types: \textit{non-existent objects}, \textit{unfaithful object attributes}, and \textit{inaccurate relationships}. Our hallucination detection pipeline addresses all three types of hallucinations. 

As depicted in Figure~\ref{fig:2_overview}, ESREAL’s hallucination detection pipeline consists of three modules: the semantic reconstruction module, the alignment module, and the scoring module. 
The semantic reconstruction module reconstructs the input image based on the generated caption. 
The alignment module aligns object tokens in the generated caption with corresponding regions in both the input image and the reconstructed one. 
Lastly, the scoring module produces scores for each hallucinated token, mainly by calculating the semantic similarity between the aligned regions. 
We elaborate on each of the components below.

\subsubsection{Semantic Reconstruction Module}\label{subsubsec:3_2_1_reconstruction_module}
The semantic reconstruction module utilizes a text-to-image model to reconstruct images from captions generated by the VLM. Specifically, we employ SDXL Turbo~\cite{sauer2023adversarial}, an enhanced version of Stable Diffusion~\cite{rombach2022highresolution} known for better prompt compliance and increased processing speed. The goal is to generate an image that faithfully reflects the provided caption. Given the inherent stochastic nature of diffusion models, the image is reconstructed four times to reduce the variance of the performance within our detection pipeline.

\subsubsection{Alignment Module}\label{subsubsec:3_2_2_alignment_module}
The alignment module, following image reconstruction, aligns specific regions between the original and reconstructed images, leveraging object tokens from the generated caption as anchors. 
This is achieved using Grounding DINO~\cite{liu2023grounding}, a proficient open object detection model capable of matching text phrases to relevant image regions.
The procedure involves two steps. 
First, Grounding DINO aligns phrases from the generated caption with corresponding areas in the reconstructed image. 
Second, it uses these aligned phrases as textual input to locate matching regions in the original image. 
This approach enables a phrase-based comparison between the original and reconstructed images.

\subsubsection{Scoring Module}\label{subsubsec:3_2_3_scoring_module}
The scoring module evaluates the aligned regions to detect and calculate penalties for hallucinated tokens based on their specific hallucination type. 
We implement distinct scoring mechanisms for each type of hallucination:
\paragraph{Non-existent Objects.}
This aspect examines whether objects in the generated caption are absent in the original image. 
The alignment module's outcome uncovers these non-existent objects.
Object phrases that are aligned in the reconstructed image, yet fail to find a corresponding image in the input image, signify hallucinated objects. 
Formally, we define a non-existent object hallucination penalty for a token \(\tau\) as 
\[
p_{\textrm{obj}}(\tau) = 
\begin{cases} 
-1 & \text{if condition } C \text{ is met},\\
0 & \text{otherwise}.
\end{cases}
\]
where \(C\) indicates that the token $\tau$ is aligned in the reconstructed image \(I_{\textrm{rec}}\) but fails to be aligned in the original image \(I_{\textrm{org}}\).

\paragraph{Unfaithful Object Attributes.}
In an image, an object region, typically represented as a bounding box, can encapsulate all the information about an object's attributes. 
Thus, if an object region in the original image greatly differs from that in the reconstructed image, it suggests that the caption accurately depicts certain attributes of the object.
For each alignment, consisting of a phrase - input image region - reconstructed image region triplet, we compute the semantic similarity between the matched regions. We employ CLIP (ViT-L-14 trained on DataComp-1B)~\cite{radford2021learning} to compute semantic similarity scores.  

Formally, an unfaithful object attribute penalty for a token \(\tau\) is defined as
\[
p_{\textrm{att}}(\tau) = \frac{\text{sim}(I_{\textrm{org}}(\tau), I_{\textrm{rec}}(\tau)) - 1}{2},
\]
where \(I_{\textrm{org}}(\tau)\) and \(I_{\textrm{rec}}(\tau)\) denotes the regions aligned to the token \(\tau\) in the original and reconstructed image, respectively. 

\paragraph{Inaccurate Relationships.}
Inaccurate relationships refer to erroneous spatial positions and interactions among objects. We focus specifically on the inaccurate spatial relationships between objects.
To evaluate inaccurate spatial relationships in generated captions, we quantify the similarity in orientation between vectors associated with pairs of objects from the original and reconstructed image.  
First, we pair object tokens ($\tau_1$, $\tau_2$) that belong to the same sentence and are linked by a positional token (e.g., left or right). 
Then, we construct two vectors $\vec{v}_{org},\vec{v}_{rec}$ by connecting the center points of the regions corresponding to these paired tokens ($\tau_1$, $\tau_2$) in the original image \(I_{\textrm{org}}(\tau_1), I_{\textrm{org}}(\tau_2)\) and the reconstructed image \(I_{\textrm{rec}}(\tau_1), I_{\textrm{rec}}(\tau_2)\), respectively. 
Lastly, we calculate the cosine similarity between these vectors as a reflection of how closely two vectors are oriented to each other.
The degree of similarity in the description of spatial relationships between \( \tau_1, \) and \(\tau_2\) is inferred from the vector orientation.
Formally, we define an inaccurate relationship penalty for the positional token \(\tau_\textrm{pos}\) as
\[
p_{\textrm{rel}}(\tau_\textrm{pos}) = \frac{\cos(\vec{v}_{\textrm{org}}(\tau_1, \tau_2), \vec{v}_{\textrm{rec}}(\tau_1, \tau_2)
) -1}{2},
\]
where \(\vec{v}_{\textrm{org}(\tau_1, \tau_2)}, \vec{v}_{\textrm{rec}(\tau_1, \tau_2)}\) are the vectors constructed from regions aligned to tokens \(\tau_1, \tau_2\) in the original and reconstructed image.
Instead of associating this score with the object token pairs \( \tau_1, \tau_2\), it is attributed to the positional token \(\tau_\textrm{pos}\) present in the same sentence.

\subsection{Fine-grained PPO}


\subsubsection{Proximal Policy Optimization}
PPO~\cite{schulman2017proximal} is a reinforcement learning method that optimizes a policy model with a reward model while preventing destructive large policy updates. 
PPO has proven to be effective for sequence generation tasks by previous works~\cite{ouyang2022training, sun2023aligning}. 
ESREAL also adopts PPO as the backbone of its training strategy. 
Within our framework, the policy \(P_{\theta}\), initialized with a pre-trained VLM, produces an action \(a_t\) corresponding to a token generated at timestep \(t\).
The state \(s_t\), which serves as the input to the policy, consists of both the sequence of prompt tokens \(x_1, ..., x_l\) and the series of previously selected actions or tokens \(a_0, ..., a_{t-1}\).
An episode terminates at timestep \(T\) either when \(T\) surpasses \(T_\text{max}\) or upon the generation of an end-of-sequence token.

\subsubsection{Fine-grained Reward}
Traditionally, PPO implementations have favored a holistic reward approach, assigning a single scalar reward at the end of the sequence. 
Contrary to this approach, we take advantage of employing a fine-grained reward system~\cite{wu2023finegrained}, which distributes rewards more densely throughout the sequence generation process. 
We extend the sentence-level reward system of ~\cite{wu2023finegrained} to allocate penalties to individual tokens associated with hallucinations. 

By leveraging the hallucination detection pipeline as the token-level reward model, we seamlessly integrate our detection mechanism into the fine-grained PPO of ESREAL.
Our hallucination detection pipeline is designed to pinpoint tokens associated with non-existent objects, unfaithful object attributes, and inaccurate relationships, and calculate type-specific penalties for each hallucinated token. 
The penalties act as fine-grained rewards in ESREAL.
This strategy enhances the granularity of reward distribution, directly targeting the tokens responsible for hallucinations. The superiority of our fine-grained reward strategy over the holistic reward approach is further validated in Table \ref{tab:4_ablation_gpt4v}.

In addition to the fine-grained penalties $p_{\textrm{obj}}(\tau)$, $p_{\textrm{att}}(\tau)$, and $p_{\textrm{rel}}(\tau)$, we incorporate a holistic reconstruction reward \(r_{\textrm{rec}}\) at the final timestep \(T\). 
This reward represents the semantic similarity between the entire original and reconstructed images.
The semantic similarity is computed via CLIP aforementioned in ~\ref{subsec:3_2_detection_pipeline}.
Formally, the holistic reconstruction reward \(r_{\textrm{rec}}\) is defined as: 
\[
r_{\textrm{rec}} = \frac{\text{sim}(I_{\textrm{org}}, I_{\textrm{rec}}) + 1}{2}.
\]
This holistic reward serves a dual purpose. 
Primarily, it functions as a regularization mechanism to prevent potential exploitation of the reward system through the omission of object and positional tokens in the generated captions. 
Also, it captures the interactions among objects, a dimension of relationship hallucinations not explicitly addressed by the fine-grained penalties.

Furthermore, we adopt an approximate KL divergence penalty to explicitly penalize divergence from the original policy, as suggested by previous studies~\cite{ziegler2020finetuning}. 

Given the fine-grained hallucination penalties, the holistic reconstruction reward, and the KL divergence penalty, we define the reward \(r_t\) for each generated token \(a_t\) as follows:

\begin{equation}
\begin{split}
    r_t = &\alpha \cdot \left(p_{\text{obj}}(a_t) + p_{\text{att}}(a_t) + p_{\text{rel}}(a_t)\right)\\ &+ (1 - \alpha) \cdot r_{\textrm{(rec, t=T)}} - \beta \log \frac{P_\theta (a_t | s_t)}{P_{\text{ref}}(a_t | s_t)},
\end{split}
\end{equation}
where \(\alpha\) denotes the hyperparameter balancing detailed reconstruction and hallucination mitigation, while \(\beta\) represents the weighting factor for the KL divergence penalty.

%% file: Figures/2_overview.tex
\begin{figure*}[!t]
\centering
\includegraphics[width=\textwidth]
{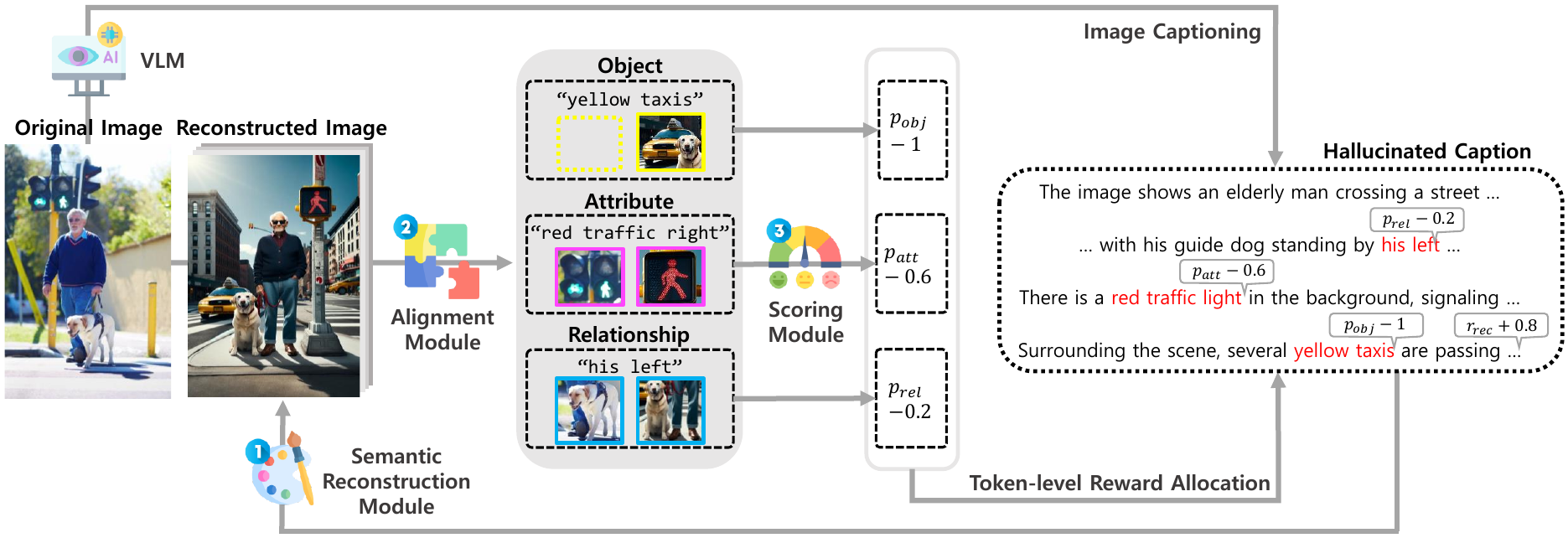}
\caption{Overview of ESREAL. 
The semantic reconstruction module, alignment module, and scoring module form a reference-free hallucination detection pipeline. 
Penalties produced by the detection pipeline for object, attribute, and relationship hallucinations are denoted as \(p_\text{obj}\),  \(p_\text{attr}\),  \(p_\text{rel}\), respectively. Penalties are allocated to the corresponding hallucinated tokens, with a holistic reconstruction reward $r_\text{rec}$ allocated at the end of the caption.
}
\label{fig:2_overview}
\end{figure*}

%% file: Sections/4_experiments.tex
\section{Experiments}\label{sec:4_experiments}

\subsection{Experimental Setup}

\noindent\textbf{Dataset.}
Since VLMs exhibit more severe hallucinations with longer sequences~\cite{wang2023evaluation}, we conduct experiments on the image paragraph captioning task. 
Image paragraph captioning is a vision-language task that requires the generation of long and detailed descriptions of the image.
Specifically, we conduct our experiments on the Stanford Image Paragraph dataset~\cite{krause2017hierarchical}, consisting of 19,561 image-caption pairs.
During the training phase, we only utilize the images, as ESREAL is an unsupervised learning framework. 
The annotated ground-truth captions in the dataset are solely used to evaluate ESREAL's performance on task-specific metrics.

\noindent\textbf{Model.}
We demonstrate ESREAL's potential as a model-agnostic approach to mitigating hallucinations by conducting experiments on three open-source VLMs:  LLaVA 1.5 (7B)~\cite{liu2023llava}, InstructBLIP FlanT5-XL (3B)~\cite{dai2023instructblip}, and mPLUG-Owl2 (7B)~\cite{ye2023mplug}.

\noindent\textbf{Implementation Details.}
We fine-tune our VLMs using the AdamW~\cite{loshchilov2019decoupled} optimizer with a learning rate of 1e-5 and a cosine annealing scheduler.
Batch sizes are set at 64 for InstructBLIP and 16 for both LLaVA and mPLUG-Owl2.
We also set the number of inference steps to 4 for SDXL Turbo~\cite{sauer2023adversarial} and employed the Long Prompt Weighting (LPW) pipeline to accommodate long captions as input.
Additional details are provided in the supplementary material.

\subsection{Evaluation Protocols and Results}

\input{Tables/1_chair_metrics}

\noindent\textbf{CHAIR Evaluation.} 
CHAIR~\cite{rohrbach2018object} is the most widely used metric to quantify the occurrence of non-existent objects in captions. 
We report the two variants of CHAIR, the sentence level \(CHAIR_s\) and object instance level \(CHAIR_i\). \(CHAIR_s\) represents the proportion of captions that embody non-existent objects. \(CHAIR_i\) calculates the ratio of non-existent objects to the total number of mentioned objects. 
We also compute Coverage~\cite{ding2023crosscodeeval}, the number of accurately mentioned objects to the total number of existent objects.

Table \ref{tab:6_chair_metrics} shows that ESREAL notably reduces hallucinated objects in captions while improving Coverage. 
These findings indicate that ESREAL not only contributes to increasing the accuracy but also enhances the level of detail by incorporating previously overlooked objects in captions.

\subsubsection{FaithScore Evaluation}

\input{Tables/2_faithscore_metrics}
In addition to CHAIR, we report results on FaithScore~\cite{jing2023faithscore}, an evaluation method capable of evaluating a broader range of hallucination types.
FaithScore recognizes descriptive sub-sentences, decomposes them to atomic facts, and verifies their factual accuracy. 
These facts fall into five categories: entity, count, color, relationship, and others. 
The process relies on powerful LLM for recognition and decomposition tasks, with Visual Entailment Models (VEMs)~\cite{liu2023llava} used for verification. 
In our case, we use GPT-3.5 Turbo as the LLM and OFA~\cite{Wang2022OFAUA} as the VEM.

In addition to the original sentence level FaithScore \(\hat{f}_\textrm{s}\), we define a sentence level FaithScore for each type of hallucination (object, attribute, and relationship) \(\hat{f}_\textrm{s, type}\). Considering the definition of each type of fact~\cite{jing2023faithscore}, we classify entity atomic fact as object, relationship as relationship, and count, color, and others as attribute. 
\(\hat{f}_\textrm{s}\) and \(\hat{f}_\textrm{s, type}\) are formulated as follows:
\[
\hat{f}_\textrm{s} = 1 - \frac{C_h}{C},\quad\hat{f}_\textrm{s, type} = 1 - \frac{C_\textrm{h, type}}{C}, \quad \textrm{type} \in \{\textrm{obj}, \textrm{att}, \textrm{rel}\}
\]
\({C_\textrm{h}}\) denotes the number of descriptive sub-sentences manifesting any hallucination, and \({C}\) signifies the total number of descriptive sub-sentences. \(C_\textrm{h, type}\) specifies the number of descriptive sub-sentences manifesting hallucination of a particular type. 

Table \ref{tab:1_faithscore_metrics} shows that ESREAL is able to reduce overall hallucinations in captioning (\(\hat{f}_\textrm{s}\)) across LLaVA, InstructBLIP, and mPLUG-Owl2. The improvements in type-wise FaithScore \(\hat{f}_\textrm{s, type}\) demonstrate that ESREAL comprehensively mitigates non-existent objects, inaccurate object attributes, and unfaithful relationships. 

\input{Tables/3_gpt4v_metrics}

\subsubsection{GPT-4V-aided Evaluation}
Recent literature~\cite{zhang2023gpt4vision, yin2023woodpecker} propose employing GPT-4V to automatically evaluate performance in vision-language tasks.
In light of this trend, we assess type-wise hallucinations with GPT-4V. 
Due to strict API limits, we conduct GPT-4V-aided evaluation on 200 randomly selected images from the test split of the Stanford Image Paragraph dataset.

Specifically, we submit the image and generated caption to GPT-4V and prompt it to list non-existent objects, unfaithful object attributes, and inaccurate relationships in the caption. 
Then, we calculate the average occurrence of each type of hallucination per caption by 
\[
\textrm{GPT-4V}_\textrm{type} = \frac{C_{type}}{C}, \quad \textrm{type} \in \{\textrm{obj}, \textrm{att}, \textrm{rel}\}
\]
where \(C_{type}\) denotes the total number of a particular type of hallucinations listed by GPT-4V and \(C\) represents the total number of generated captions. 
Table \ref{tab:2_gpt4v_metrics} demonstrates that ESREAL leads to a substantial reduction in each type of hallucination across all three VLMs.
More details on the prompt and output of GPT-4V-aided evaluation are provided in the supplementary material.

\input{Tables/4_task_metrics}

\subsubsection{Task-Specific Evaluation}
We report three conventional image paragraph captioning metrics, CIDEr~\cite{vedantam2015cider}, ROUGE-L~\cite{lin-2004-rouge}, and BLEU~\cite{papineni-etal-2002-bleu} as well as hallucination metrics. This is to demonstrate that ESREAL preserves the generative capabilities of VLMs while mitigating hallucinations. 
Table \ref{tab:3_task_metrics} reveals that ESREAL's performance is either comparable to or exceeds that of the baseline models, especially noting a marked improvement in CIDEr scores.
These findings suggest that ESREAL can be integrated with ongoing efforts to improve generative performance.

%% file: Tables/1_chair_metrics.tex
\begin{table*}[!t]
    \begin{center}
    \scriptsize
    \caption{CHAIR evaluation results.}
    
    \begin{tabular}{c|c|ccc}
        \toprule
        
        \multirow{2}{*}{Model} & \multirow{2}{*}{Method} & \multicolumn{3}{c}{CHAIR} \\
        & & $CHAIR_s$ ($\downarrow$) & $CHAIR_i$ ($\downarrow$) & Coverage ($\uparrow$) \\
        \midrule
        
        \multirow{2}{*}{\textit{LLaVA}}
          & Baseline & 0.64 & 0.18 & \textbf{0.45} \\
          & ESREAL & \textbf{0.43} & \textbf{0.12} & 0.42 \\
          
        \midrule

         \multirow{2}{*}{\textit{InstructBLIP}}
          & Baseline & 0.48 & 0.16 & \textbf{0.36} \\
          & ESREAL & \textbf{0.35} & \textbf{0.12} & 0.34 \\

        \midrule
        
        \multirow{2}{*}{\textit{mPLUG-Owl2}}
          & Baseline & 0.67 & 0.20 & \textbf{0.44} \\
          & ESREAL & \textbf{0.62} & \textbf{0.18} & \textbf{0.44} \\
          
        \bottomrule
    \end{tabular}%
    
    \label{tab:6_chair_metrics}
    \end{center}
\end{table*}

%% file: Tables/2_faithscore_metrics.tex
\begin{table*}[!t]
    \caption{FaithScore evaluation results}

    \begin{center}
    \scriptsize 
    
    \begin{tabular}{c|c|cccc}
        \toprule
        
        \multirow{2}{*}{Model} & \multirow{2}{*}{Method} & \multicolumn{4}{c}{FaithScore ($\uparrow$)} \\
        & & $\hat{f}_\textrm{s}$& $\hat{f}_\textrm{s,obj}$ & $\hat{f}_\textrm{s,att}$  & $\hat{f}_\textrm{s,rel}$ \\
        \midrule
        
        \multirow{2}{*}{\textit{LLaVA}}
          & Baseline & 0.7401 & 0.8346 & 0.9105 & 0.8335 \\
          & ESREAL & \textbf{0.7846} & \textbf{0.8649} & \textbf{0.9301} & \textbf{0.8810}\\
          
        \midrule

        \multirow{2}{*}{\textit{InstructBLIP}}
          & Baseline & 0.7113 & 0.8066 & 0.9003 & 0.8304 \\
          & ESREAL & \textbf{0.7834} & \textbf{0.8459} & \textbf{0.9197} & \textbf{0.8763} \\

        \midrule
        
        \multirow{2}{*}{\textit{mPLUG-Owl2}}
          & Baseline & 0.7171 & \textbf{0.8246} & 0.8924 & 0.8043 \\
          & ESREAL & \textbf{0.7202} & 0.8238 & \textbf{0.8938} & \textbf{0.8068} \\
          
        \bottomrule
    \end{tabular}%
    
    \label{tab:1_faithscore_metrics}
    \end{center}
\end{table*}

%% file: Tables/3_gpt4v_metrics.tex
\begin{table*}[!t]
\caption{GPT-4V-aided evaluation results.}

\begin{center}
\scriptsize

\begin{tabular}{c|c|cccc}
    \toprule
    
    \multirow{2}{*}{Model} & \multirow{2}{*}{Method} & \multicolumn{4}{c}{\# Hallucinations per Caption ($\downarrow$)} \\
    & & Object & Attribute & Relationship & Total \\
      
    \midrule

     \multirow{2}{*}{\textit{LLaVA}}
      & Baseline & 1.07 & 0.12 & 0.87 & 2.06 \\
      & ESREAL & \textbf{0.66} & \textbf{0.08} & \textbf{0.78} & \textbf{1.52} \\
      
    \midrule
    
     \multirow{2}{*}{\textit{InstructBLIP}}
      & Baseline & 1.23 & 0.14 & 1.05 & 2.42 \\
      & ESREAL & \textbf{0.80} & \textbf{0.06} & \textbf{0.64} & \textbf{1.49} \\
      
    \midrule
    
     \multirow{2}{*}{\textit{mPLUG-Owl2}}
      & Baseline & 1.25 & 0.14 & 1.33 & 2.51 \\
      & ESREAL & \textbf{1.21} & \textbf{0.11} & \textbf{1.19} & \textbf{2.32} \\
      
    \bottomrule
\end{tabular}%

\label{tab:2_gpt4v_metrics}
\end{center}
\end{table*}

%% file: Tables/4_task_metrics.tex
\begin{table*}[!t]
\caption{Image paragraph captioning task evaluation results.}

\begin{center}
\scriptsize

\begin{tabular}{c|c|c|c|cccc}
    \toprule
    
    \multirow{2}{*}{Model} &
    \multirow{2}{*}{Method} &
    \multirow{2}{*}{CIDEr ($\uparrow$)} &
    \multirow{2}{*}{ROUGE-L ($\uparrow$)} &
    \multicolumn{4}{c}{BLEU ($\uparrow$)} \\
    & & & & BLEU-1 & BLEU-2 & BLEU-3 & BLEU-4  \\
    
    \midrule
    
    \multirow{2}{*}{\textit{LLaVA}}
      & Baseline & 4.52 & 
      21.55 & 26.81 & 13.42 & 6.31 & 3.08  \\
      & ESREAL & \textbf{9.10} & 
      \textbf{23.02} & \textbf{31.56} & \textbf{16.09} & \textbf{7.67} & \textbf{3.72} \\

    \midrule

    \multirow{2}{*}{\textit{InstructBLIP}}
      & Baseline & 5.99 & 
      \textbf{22.68} & 26.79 & 13.68 & 6.76 & 3.51  \\
      & ESREAL & \textbf{7.85} & 
      21.90 & \textbf{29.79} & \textbf{14.82} & \textbf{6.99} & \textbf{3.56}  \\
      
    \midrule
    
    \multirow{2}{*}{\textit{mPLUG-Owl2}}
      & Baseline & 3.71 & 
      \textbf{21.46} & \textbf{26.64} & \textbf{13.29} & 6.13 & 2.94  \\
      & ESREAL & \textbf{3.99} & 
      \textbf{21.46} & 26.60 & 13.26 & \textbf{6.14} & \textbf{2.96}  \\
      
    \bottomrule
\end{tabular}%

\label{tab:3_task_metrics}
\end{center}
\end{table*}

%% file: Sections/5_analysis.tex
\section{Analysis}



\subsection{Ablation Study}

\input{Tables/5_ablation_gpt4v}
To demonstrate the impact of individual components of our reward design, we conduct ablation experiments on each term in the reward formulation.
We conduct four experiments, each omitting one of the hallucination penalties $p_{\textrm{obj}}(t)$, $p_{\textrm{att}}(t)$, $p_{\textrm{rel}}(t)$, or the holistic reconstruction reward \(r_{\textrm{rec}}\).
We utilize the GPT-4V-aided evaluation aforementioned in Section~\ref{sec:4_experiments} for hallucination assessment.
Table~\ref{tab:4_ablation_gpt4v} shows that the removal of each type of hallucination penalty results in a notable increase in the corresponding type of hallucination. 
This indicates that these penalties effectively target and reduce the occurrences of their respective hallucination types.

Removing the holistic reconstruction reward $r_\textrm{rec}$ leads to a significant decrease in the CIDEr score by 51\% (from 7.85 to 3.83). We observe that, in the absence of the reward term, VLMs resort to reward hacking by generating minimal tokens to evade penalties. This finding underscores the role of $r_\textrm{rec}$ as a crucial regularization mechanism that deters such behavior.
In addition, the occurrence of inaccurate relationships significantly increases without $r_\textrm{rec}$. 
Such observation can be attributed to the role of $r_\textrm{rec}$ in denoting object interactions overlooked by $p_{\textrm{rel}}(t)$, highlighting its importance in mitigating relationship hallucinations. 

In addition, we perform an ablation study to examine the effects of token-level penalty allocation, employing only the holistic reconstruction reward $r_\textrm{rec}$. The resulting hallucination metrics reveal that the fine-grained penalties play a critical role in ESREAL's ability to mitigate hallucinations. 


\input{Figures/3_reward_analysis}

\subsection{Stability Analysis}
ESREAL integrates various models: SDXL Turbo~\cite{sauer2023adversarial} in the semantic reconstruction module, Grounding DINO\cite{liu2023grounding} in the alignment module, and the CLIP encoder~\cite{radford2021learning} in the scoring module. Errors in these models are inevitable but may raise concerns about ESREAL's stability. This section aims to examine ESREAL's stability beyond its experimental success.
\subsubsection{Win Rate of Rewards}
The stability of ESREAL depends on the consistency of rewards generated by its hallucination detection pipeline. A stable reward system would impose lower penalties on accurate captions and higher penalties on hallucinated ones. 
Upon this observation, we introduce the Win Rate of Rewards, quantifying the proportion of instances in which the reward model correctly assigns greater penalties to hallucinated captions over accurate ones. 
Utilizing GPT-4, we generate hallucinated caption pairs for each annotated caption in the Stanford Image Paragraph dataset, extending prior research~\cite{jiang2024hallucination, liu2023mitigating}. Figure \ref{fig:3_reward_analysis} (a) demonstrates that our reward model achieves an overall Win Rate of 78\%, with similar results across different hallucination penalty types, indicating a stable distribution of rewards.
\subsubsection{Aggregating Rewards}
The stochastic properties of the diffusion model may be advantageous for generating diverse images, but there is a risk of compromising the stability of reward production in our approach. 
To address this concern, our framework aggregates the scores of multiple reconstructed images. Figure~\ref{fig:3_reward_analysis} (b) illustrates the correlation between the win rate and the number of reconstructed images. 
The analysis results show that employing multiple reconstructed images improves the stability of the reward.

\subsection{Case Analysis}

\input{Figures/4_case_analysis}

\subsubsection{Reward Allocation}
In Figure~\ref{fig:4_case_analysis}, we illustrate two instances of token-level penalty allocation on hallucinations. The ESREAL reward model assigns a -1 object hallucination score to the non-existent ``tennis racket'' in the upper image. In the lower image, it appropriately penalizes the inaccurate spatial relationship ``left'' with a -0.8 score.

\input{Figures/5_case_analysis}

\subsubsection{Hallucination Mitigation}
We demonstrate two instances where captions created by LLaVA, before and after the integration of ESREAL, show significant improvements in Figure~\ref{fig:5_case_analysis2}. 
Specifically, after implementing ESREAL, LLaVA no longer fabricates objects such as ``clock'' or ``another tree''. 
Moreover, spatial inaccuracies like ``book placed on the bed'' and ``fire hydrant near the center'' are accurately depicted in the captions after ESREAL integration. Additionally, erroneous attributes, such as ``three cats,'' are corrected to ``two cats'' post-ESREAL application.

\subsection{Limitations}

Despite the clear advantages of being an unsupervised yet effective method to mitigate hallucinations in VLMs, we acknowledge certain limitations of ESREAL. 
Firstly, ESREAL necessitates semantic reconstruction from the VLM's response, requiring attention to the entire image. 
This inherent characteristic restricts its applicability solely to image captioning tasks. 
Secondly, the efficacy of ESREAL's detection pipeline depends on the integrated performance of a text-to-image model, a grounding model, and a similarity model. 
An error in any of these components could propagate, impacting the accuracy of the entire system. 
We anticipate that the reliability of ESREAL will enhance with technological advancements in the related fields of study. 
Thirdly, our approach to penalizing inaccurate relationship hallucinations primarily focuses on the spatial positions between objects. 
This limitation reflects the challenges identified in previous works, where quantifying relationship hallucinations remains a complex issue. 
We acknowledge the necessity to penalize a broader range of relationship hallucinations.

%% file: Tables/5_ablation_gpt4v.tex
\begin{table*}[!t]
    \caption{Ablation study.}
    \begin{center}
    \scriptsize
    
    \begin{tabular}{c|c|cccc}
        \toprule
        
        \multirow{2}{*}{Model} & \multicolumn{1}{c|}{\multirow{2}{*}{Method}} & \multicolumn{4}{c}{\# Hallucinations per Caption ($\downarrow$)} \\
        & & Object  
        & Attribute 
        & Relationship  
        & Total \\
        
        \midrule
        
         \multirow{7}{*}{\textit{InstructBLIP}}
          & Baseline & 1.23 & 0.14 & 1.05 & 2.42 \\ \cmidrule{2-6}
          & ESREAL & 0.80 & 0.06 & 0.64 & 1.49 \\ \cmidrule{2-6}
          & ESREAL w/o $p_{\textrm{obj}}$ & 1.18 & 0.14 & 0.92 & 2.24 \\
          & ESREAL w/o $p_{\textrm{att}}$ & 0.93 & 0.09 & 0.59 & 1.61 \\
          & ESREAL w/o $p_{\textrm{rel}}$ & 0.85 & 0.07 & 1.28 & 2.21 \\
          & ESREAL w/o $r_{\textrm{rec}}$ & 0.68 & 0.04 & 1.40 & 2.12 \\
          & ESREAL w/o $p_{\textrm{obj}}$, $p_{\textrm{att}}$, $p_{\textrm{rel}}$ & 1.15 & 0.10 & 1.01 & 2.26 \\
          
        \bottomrule
    \end{tabular}%
    
    \label{tab:4_ablation_gpt4v}
    \end{center}
\end{table*}

%% file: Figures/3_reward_analysis.tex
\begin{figure}[!t]
\centering
\includegraphics[width=0.85\textwidth]{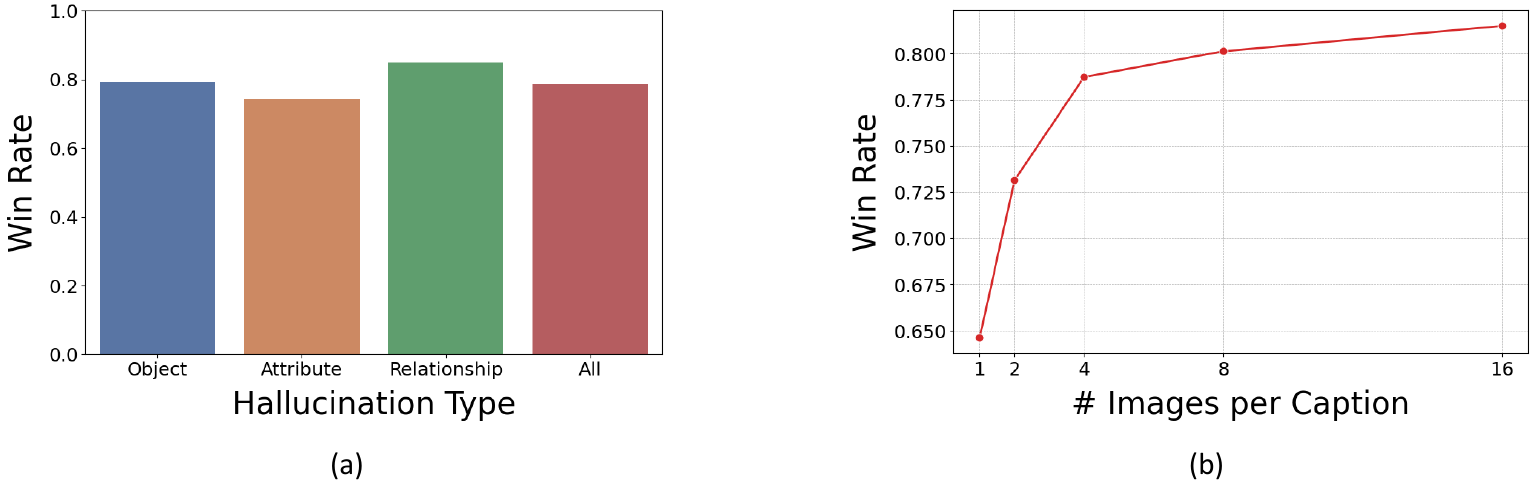}
\caption{Stability analysis. (a) a bar chart illustrating the win rates associated with different hallucination types. 
(b) a positive correlation between win rates and the number of images generated by the reconstruction module per caption.}
\label{fig:3_reward_analysis}
\end{figure}

%% file: Figures/4_case_analysis.tex
\begin{figure*}[!t]
\centering
\includegraphics[width=\textwidth]{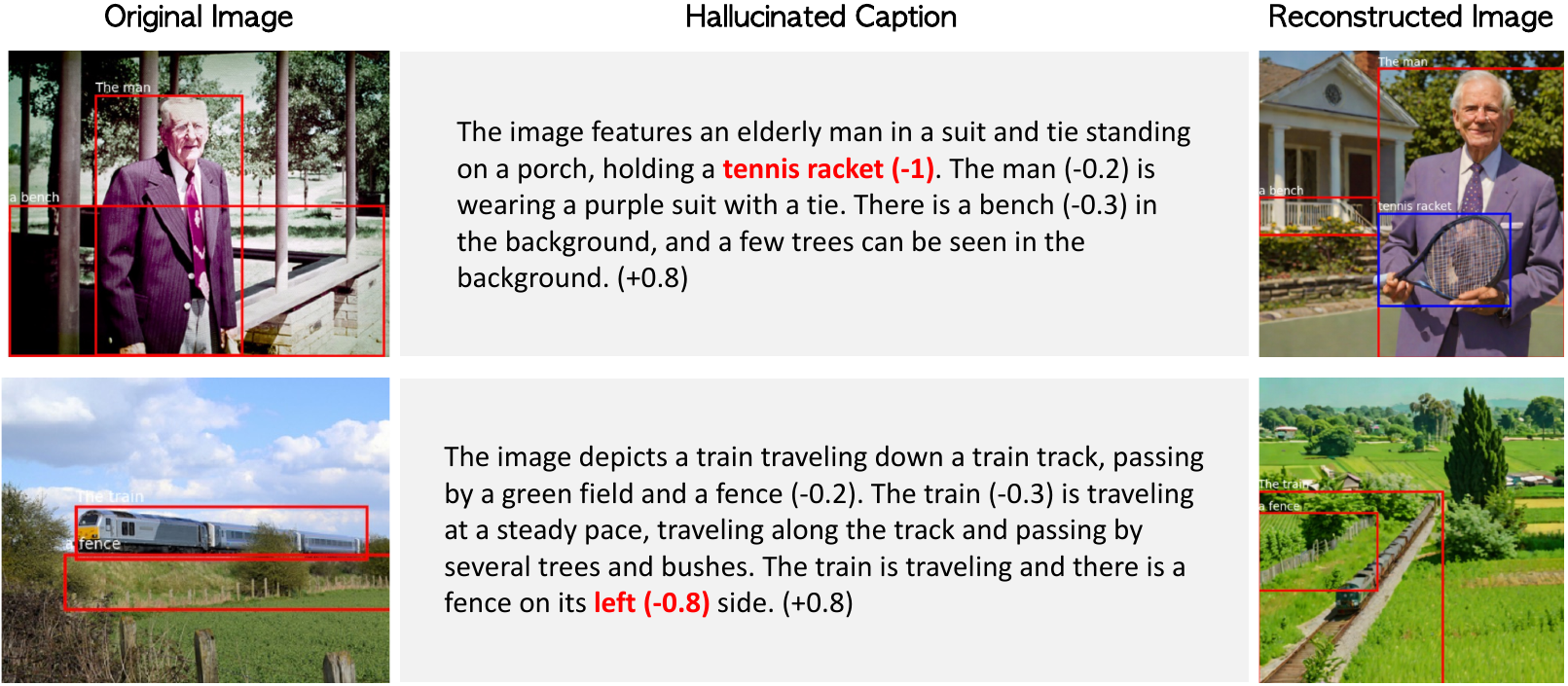}
\caption{A case study of reward allocation on hallucinated captions generated by LLaVA.}
\label{fig:4_case_analysis}
\end{figure*}

%% file: Figures/5_case_analysis.tex
\begin{figure*}[!t]
\centering
\includegraphics[width=\textwidth]{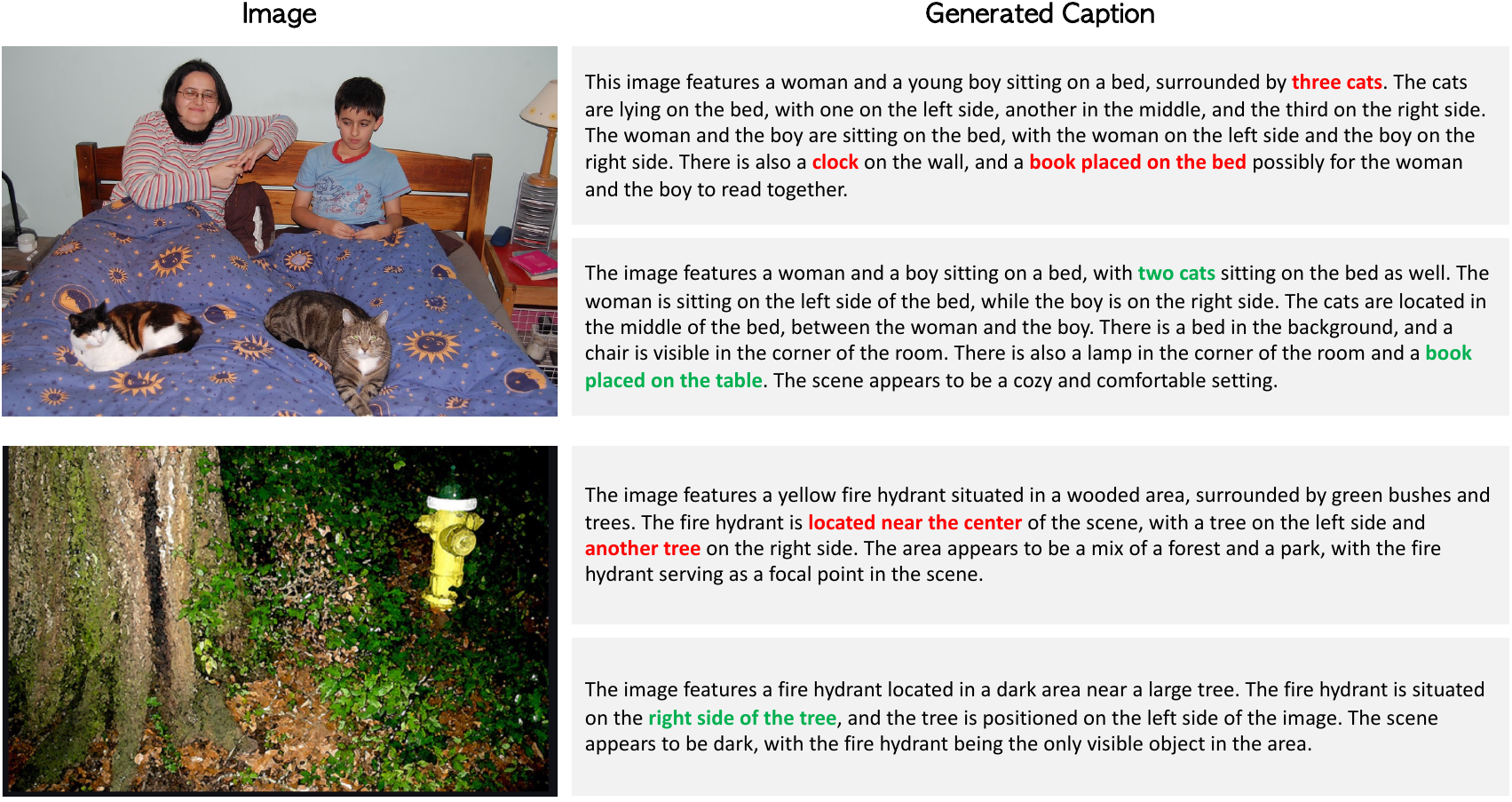}
\caption{A case study comparing the captions generated before and after training LLaVA with ESREAL.}
\label{fig:5_case_analysis2}
\end{figure*}

%% file: Sections/6_conclusion.tex
\section{Conclusion}
In this paper, we introduce ESREAL, an unsupervised learning framework designed to address hallucinations in VLMs during image captioning tasks by leveraging semantic reconstruction. 
We demonstrate that by providing fine-grained negative feedback to hallucinated tokens, VLMs can effectively learn to suppress hallucinatory content without compromising their generative capabilities. 
ESREAL exhibits two notable characteristics: 
(1) a novel reference-free hallucination detection pipeline that accurately identifies hallucinated tokens along with their specific types, and (2) a mechanism for assigning fine-grained rewards to these tokens. 
Our experiments, conducted with three pre-trained VLMs, reveal that ESREAL consistently reduces hallucinations as evaluated by FaithScore and GPT-4V, while maintaining or improving performance on key task metrics such as CIDEr, ROUGE-L, and BLEU.
Our research underscores the potential of unsupervised approaches in enhancing the reliability of multimodal models, particularly in scenarios where paired data across different modalities is scarce. 
We make our implementation of ESREAL publicly available, aiming to contribute to the development of more dependable and robust AI systems across diverse domains and applications.

%% file: Sections/aknowledgements.tex
\section*{Acknowledgements}
We are grateful to VESSL AI for providing the compute resources used for the experiments in this work.

%% file: Sections/A_appendix.tex
\clearpage
\appendix

\section{Rationale for Semantic Reconstruction}
When humans identify hallucinations within a generated caption, a pivotal step would involve comparing segments of the input image with the corresponding parts of the generated caption. 
For instance, to determine whether the statement ‘A small white dog is sleeping.’ contains hallucinations, one would first seek to locate the dog in the image, and then compare its features with the descriptions in the statement.  

An ideal hallucination detection scheme would replicate this process. 
However, despite the technological advancements, the accurately binding portions of text to corresponding object regions in an image remains a challenging endeavor due to a difference in how objects are encapsulated in visual and textual data.
Identifying which portion of the image corresponds to a specific object is relatively straight forward, because locality is clear in visual data. 
Typically, the pixels forming an object are confined to a specific region, allowing straightforward encapsulation, such as within a rectangular bounding box. 

In contrast, in a paragraph, defining segments relevant to a specific object is considerably more complex. 
Unlike in images, the concept of object locality is not as pronounced in textual data. 
Descriptions of objects may be dispersed throughout the paragraph, making the search for segments of text pertinent to a specific object a difficult task. 
Even if the search is successful, another demanding aspect is reorganizing the segments to maintain the original semantic context. 
The segments of text within a paragraph are intricately connected to the surrounding phrases. 
Consequently, simple concatenation of the located object descriptions does not guarantee accurate interpretation, as their meaning can alter significantly without surrounding context. 
This necessitates a careful reorganization of the text to maintain the intended meaning, presenting another complex language task.

To surmount these difficulties, we propose a innovative solution: translating the text generated by the model to the visual domain. 
By converting model generated paragraphs into images, binding regions pertinent to a specific object becomes a easier task. 
Then, a comparative analysis on these aligned regions reveals hallucinations.  
Our approach in identifying hallucinations through semantic reconstruction and training models to mitigate hallucinations proves to be an effective approach in tackling hallucinations. A significant benefit of our method is that it is entirely unsupervised, eliminating the need for extra data beyond the images themselves.

\section{Training Details}\label{sec:appendix_details}
\subsection{Dataset Statistics}
The Stanford Image Paragraph Dataset~\cite{krause2017hierarchical} contains a total of 19561 images with 14579 train images, 2490 validation images, and 2492 test images. 
The length of the corresponding detailed captions are on average 69.69 BERT tokens, 60.88 words, and 309.71 characters. 

\subsection{Training Hyperparameters}
\input{Tables/6_hyperparameters}

We describe the hyperparameters for ESREAL as follows: The sequence length is configured to be 256, with a total of 800 training steps. The batch size is set at 64 for InstructBLIP, while for LLaVA and mPLUG-Owl2, it is 16. 
The optimization is performed using the AdamW Optimizer, which has betas set to (0.9, 0.95), epsilon at 1.0e-8, and a weight decay of 1.0e-6. Additionally, a cosine annealing scheduler is employed.
For Proximal Policy Optimization (PPO), we use 4 epochs, with gamma set to 1 and lambda at 0.95. The clipping range for PPO is 0.2, similar to the value function clipping range, which is also 0.2. 
The coefficient for the value function is set at 1, and the clipping range for rewards is established at 10. The reward function is weighted with alpha at 0.8 and beta at 0.001.

\subsection{Trainable Components in VLMs}
\input{Tables/7_frozen_modules}
Due to the substantial number of parameters in contemporary VLMs, it is standard practice to keep some components of VLMs frozen while allowing others to be trainable~\cite{li2023blip2, liu2023visual}. 
We also adopt this approach when applying ESREAL on LLaVA 1.5, InstructBLIP, and mPLUG-Owl2.
In determining the trainable components, we generally comply with the the training strategies specific to each VLM, with an exception to InstructBLIP, where we additionally train the LLM. 
For all fine-tuning processes, we employ Low-Rank Adaptation for large language models (LoRA)~\cite{hu2021lora}. 
The trainable modules for each VLM are detailed in Table~\ref{tab:7_frozen_modules}. 

\subsection{Optimization of Reward Model}
ESREAL employs several large-scale models in its reward mechanism, necessitating efficient operation within acceptable memory and time constraints. 
Proximal Policy Optimization (PPO) offers the advantage of not requiring a differentiable reward model, allowing us to improve throughput and latency by utilizing a NVIDIA Triton Inference Server. 

We deploy key components of the reward model —SDXL Turbo (reconstruction module), Grounding DINO (alignment module), and the CLIP encoder (scoring module)— on this server. 
The critical bottleneck of our training procedure, the reconstruction and alignment modules, are distributed across multiple GPUs.
The optimized architecture and Triton Inference Server code can be found on our GitHub. 

This optimization enables a single reward computation cycle, which calculates the aggregated penalty for four reconstructed images generated from one caption, to complete in 3.27 seconds on a single GPU.

\subsection{Module-specific Details}
\subsubsection{Positional Tokens}
Positional tokens are tokens that denote spatial relationships. Specifically, we view [`left', `right', `top', `bottom', `center', `middle', `above', `below', `inside', `outside', `front', `behind', `upward', `downward', `up', `down', `inward', `outward', `over', `under'] as positional tokens. 
\subsubsection{Case Study}
\input{Figures/6_module_details}
In Figure~\ref{fig:6_module_details}, we demonstrate a case study of our alignment module and scoring module.

The alignment module first aligns the caption with the reconstructed image. As a result, the phrases - elderly man, guide dog, red traffic light, and yellow taxis - are matched to their corresponding regions in the reconstructed image. 
Then, the concatenated phrase `elderly man, guide dog, red traffic light, yellow taxis' is aligned to the original image. As a result, only `yellow taxis' remains unaligned while the other phrases find matching regions in the original image.
Using the object phrases as an anchor, regions from the reconstructed image and the original image are aligned. 

The scoring module produces hallucination penalties for non-existent objects, unfaithful attributes and inaccurate relationships. 
The unaligned `yellow taxis' reveals a hallucinated object. Accordingly, a non-existent object penalty of -1 is allocated to `taxis'.
The aligned regions for `red traffic light' are passed to the CLIP encoder and receive a low similarity score of -0.6, due to the misrepresentation of a green light as red. This unfaithful object attribute penalty is allocated to `light'.
The object phrases `elderly man' and `guide dog' are paired because they are linked within the same sentence through a positional token `left'. Two vectors are constructed by connecting the center points of the aligned regions for `elderly man' and `guide dog' in the original and reconstructed image. The orientation of these two vectors, assessed by the cosine of their angle, receives a -0.2 penalty, indicating an inaccurate spatial relationship. The penalty is allocated to the positional token `left'. 





\section{ESREAL on COCO}
We conduct our main experiments on the Stanford Image Paragraph Dataset~\cite{krause2017hierarchical}. The dataset offers the benefit of being able to evaluate ESREAL on standard captioning metrics due to its human-annotated captions. 
Therefore, we fine-tune VLMs with the training images of the dataset and evaluate their performance on its test split.

However, since ESREAL is an unsupervised method, one can utilize any image dataset, including those without annotated detailed captions. To showcase ESREAL's versatility across different datasets, we conduct experiments using the MS COCO 2014 dataset~\cite{lin2015microsoft}. 
Specifically, we fine-tune InstructBLIP FlanT5-XL on 32,000 images sampled from the COCO 2014 Karpathy train split and evaluated its performance on the test split of the Stanford Image Paragraph Dataset.
The evaluation of ESREAL on COCO aligns with the findings from the main experiment, demonstrating its ability to mitigate hallucinations without impairing the generative performance of VLMs.

\subsection{Hallucination Evaluation}
\input{Tables/8_chair_coco}
We report the CHAIR metric to evaluate the generated captions in terms of hallucinations.
Table~\ref{tab:8_chair_metrics_coco} shows that ESREAL effectively decreased both the sentence level and object instance level CHAIR scores while maintaining Coverage, indicating successful mitigation of hallucinations. 

\subsection{Captioning Evaluation}
\input{Tables/9_task_coco}
We also report the standard captioning metric to evaluate the quality of the generated captions.
Table~\ref{tab:9_task_metrics} shows that ESREAL improves captioning performance beyond the baseline InstructBLIP, notably increasing the CIDEr score.

\section{Qualitative Examples}
\input{Figures/7_qualitative_example1}
\input{Figures/8_qualitative_example2}
\input{Figures/9_qualitative_example3}
We provide additional qualitative examples to demonstrate the effectiveness of ESREAL.
Figures \ref{fig:7_qualitative_1}, \ref{fig:8_qualitative_2}, \ref{fig:9_qualitative_3} show that ESREAL effectively mitigates hallucinations in captions across different types of hallucinations.

In figure \ref{fig:7_qualitative_1}, the caption generated by baseline LLaVA contains non-existent objects `dining table visible in the background' and `another chair placed further away'. There is also an incorrect attribute `two chairs'. After fine-tuning with ESREAL, LLaVA is able to correctly identify the `blue background' and `a chair'. 

Figure \ref{fig:8_qualitative_2} shows an example of an inaccurate spatial relationship in the original caption, `some sheep standing on top of it'. Post ESREAL, the caption accurately depicts the spatial relationship between the sheep and hay pile as `some closer to the hay and others further away'. Additionally, it is able to capture the `few sheep located in the background'. 

The baseline caption in figure \ref{fig:9_qualitative_3} contains a hallucinated object `several cars'. LLaVA with ESREAL, however, does not fabricate non-existent cars anymore. Moreover, it is able to identify additional attributes of the wagon such as `old-fashioned', `covered with a canvas top' and the `dirt road'. 

\section{GPT Prompting Details}
\subsection{GPT-4V-aided evaluation}
\input{Figures/10_gpt4v_prompt}
\input{Figures/11_gpt4v_output}
Figure \ref{fig:10_gpt4v_prompt} provides the prompt we used for GPT-4V-aided evaluation.
Figure \ref{fig:11_gpt4v_output} illustrates an example output of GPT-4V-aided evaluation.
\subsection{Generating Hallucinated Captions for Win Rate Analysis}
\input{Figures/11_winrate_object}
\input{Figures/12_winrate_attribute}
\input{Figures/13_winrate_relation}
\input{Figures/14_winrate_output}
Figures \ref{fig:11_winrate_object}, \ref{fig:12_winrate_attribute}, \ref{fig:13_winrate_relations} show the prompts we used to generate hallucinated captions containing non-existent objects, unfaithful object attributes, and inaccurate relationships, respectively.
The hallucinated captions were used to calculate the Win Rate of rewards in the Stability Analysis section.
Figure \ref{fig:14_winrate_output} provides an example of the hallucinated captions generated by GPT-4.

%% file: Tables/6_hyperparameters.tex
\begin{table}[!ht]
\begin{center}
\small
\caption{Training hyperparameters.}
\begin{tabular}{lccc}
\toprule
HPs & InstructBLIP & LLaVA & mPLUG-Owl2 \\ \midrule
Training steps & 800 & 800 & 800 \\
Batch size & 64 & 16 & 16 \\
Learning rate & 1e-5 & 1e-5 & 1e-5 \\ \bottomrule
\end{tabular} %
\label{tab:6_hyperparameters}
\end{center}
\small Note: All models use AdamW optimizer with \( \epsilon = 1e-6 \), \( \beta = (0.9, 0.95) \), and weight decay of \( 1e-6 \).
\end{table}

%% file: Tables/7_frozen_modules.tex
\begin{table}[!ht]
\begin{center}
\small
\caption{Trainable VLM components.}
\begin{tabular}{lccc}
\toprule
Components & LLaVA & InstructBLIP & mPLUG-Owl2 \\ \midrule
Visual encoder & x & x & o \\
Modality connection module & o & o & o \\
LLM & o & o & o \\ \bottomrule
\end{tabular} %
\label{tab:7_frozen_modules}
\end{center}
\small Note: `o' denotes trainable components, and `x' denotes frozen components.
\end{table}

%% file: Figures/6_module_details.tex
\begin{figure*}[!t]
\centering
\includegraphics[width=\textwidth]{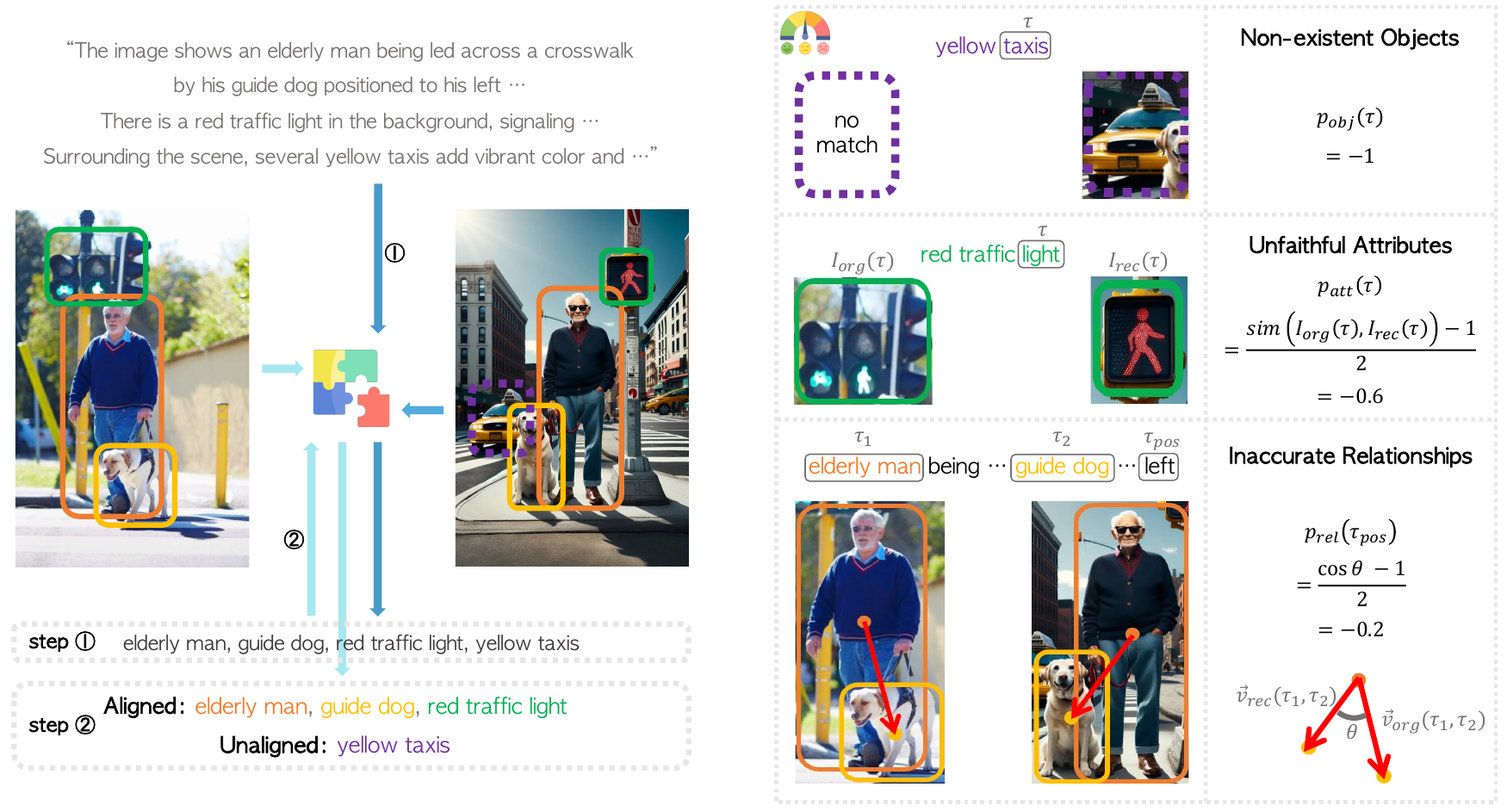}
\caption{Detailed illustration of the alignment module and the scoring module.}
\label{fig:6_module_details}
\end{figure*}

%% file: Tables/8_chair_coco.tex
\begin{table*}[!t]
    \begin{center}
    \scriptsize
    \caption{CHAIR evaluation results on COCO.}
    
    \begin{tabular}{c|c|ccc}
        \toprule
        
        \multirow{2}{*}{Model} & \multirow{2}{*}{Method} & \multicolumn{3}{c}{CHAIR} \\
        & & $CHAIR_s$ ($\downarrow$) & $CHAIR_i$ ($\downarrow$) & Coverage ($\uparrow$) \\
        \midrule

         \multirow{2}{*}{\textit{InstructBLIP}}
          & Baseline & 0.48 & 0.16 & \textbf{0.36} \\
          & ESREAL & \textbf{0.45} & \textbf{0.15} & \textbf{0.36} \\
          
        \bottomrule
    \end{tabular}%
    
    \label{tab:8_chair_metrics_coco}
    \end{center}
\end{table*}

%% file: Tables/9_task_coco.tex
\begin{table*}[!t]
\caption{Image paragraph captioning task evaluation results on COCO.}

\begin{center}
\scriptsize

\begin{tabular}{c|c|c|c|cccc}
    \toprule
    
    \multirow{2}{*}{Model} &
    \multirow{2}{*}{Method} &
    \multirow{2}{*}{CIDEr ($\uparrow$)} &
    \multirow{2}{*}{ROUGE-L ($\uparrow$)} &
    \multicolumn{4}{c}{BLEU ($\uparrow$)} \\
    & & & & BLEU-1 & BLEU-2 & BLEU-3 & BLEU-4  \\
    
    \midrule

    \multirow{2}{*}{\textit{InstructBLIP}}
      & Baseline & 5.99 & 
      \textbf{22.68} & 26.79 & 13.68 & 6.76 & 3.51  \\
      & ESREAL & \textbf{8.31} & 
      \textbf{22.68} & \textbf{30.43} & \textbf{15.48} & \textbf{7.69} & \textbf{4.06}  \\
      
    \bottomrule
\end{tabular}%

\label{tab:9_task_metrics}
\end{center}
\end{table*}

%% file: Figures/7_qualitative_example1.tex
\begin{figure*}[!t]
\centering
\includegraphics[width=0.7\textwidth]{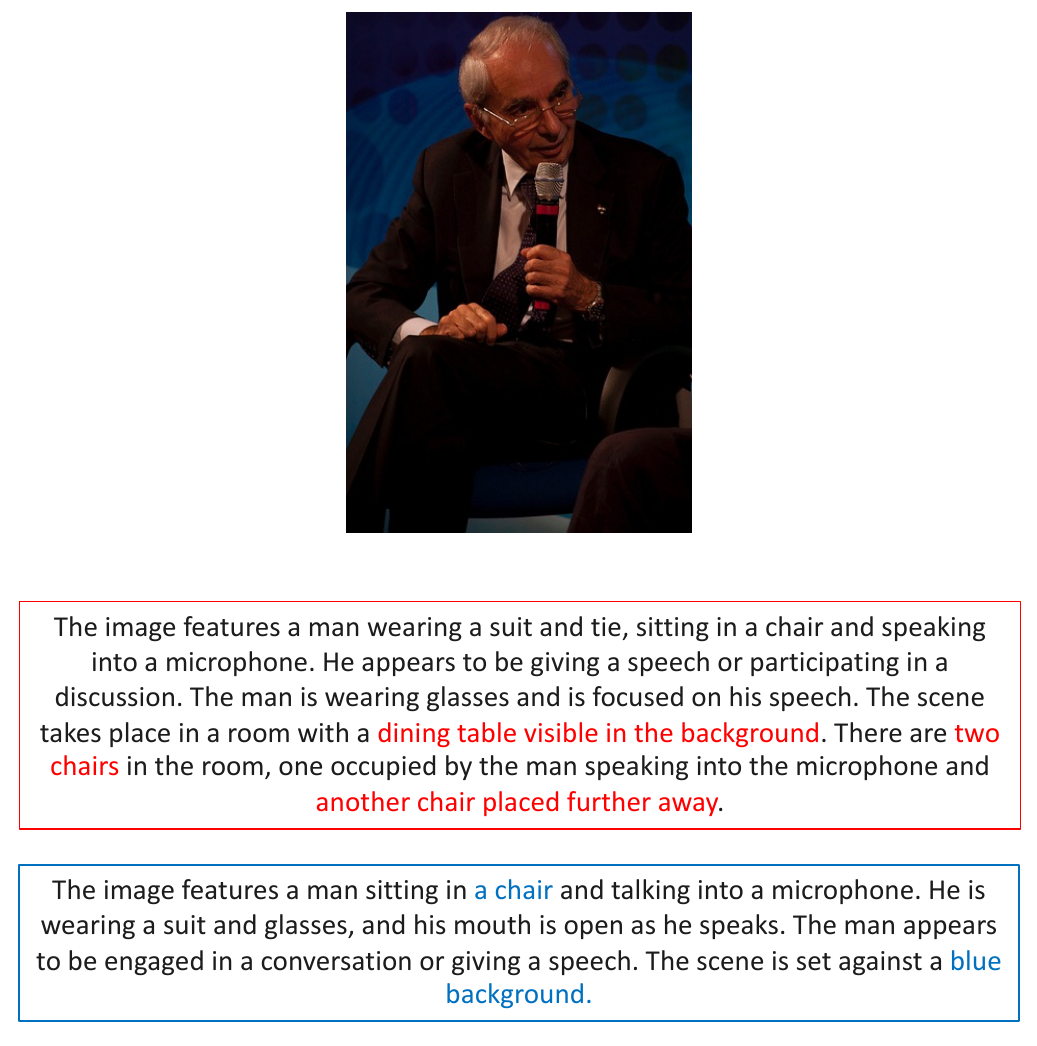}
\caption{Captions generated by LLaVA before and after ESREAL.}
\label{fig:7_qualitative_1}
\end{figure*}

%% file: Figures/8_qualitative_example2.tex
\begin{figure*}[!t]
\centering
\includegraphics[width=0.7\textwidth]{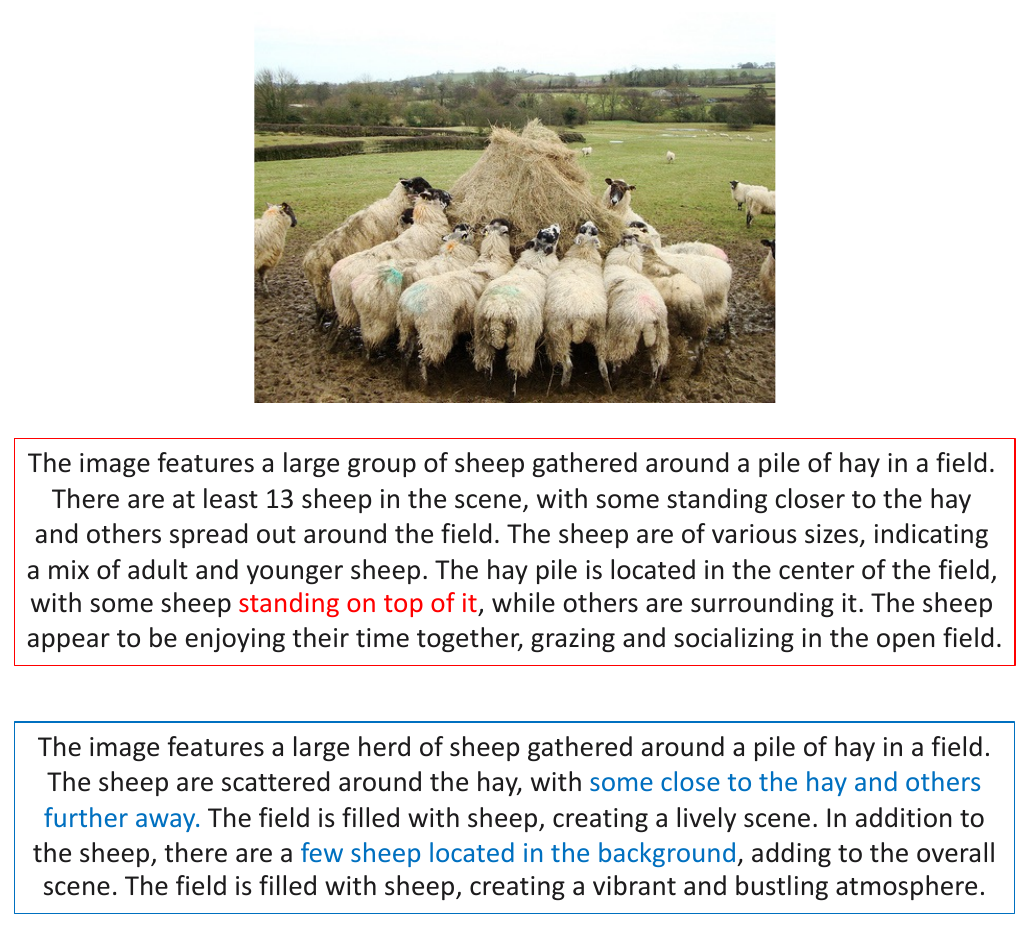}
\caption{Captions generated by LLaVA before and after ESREAL.}
\label{fig:8_qualitative_2}
\end{figure*}

%% file: Figures/9_qualitative_example3.tex
\begin{figure*}[!t]
\centering
\includegraphics[width=0.7\textwidth]{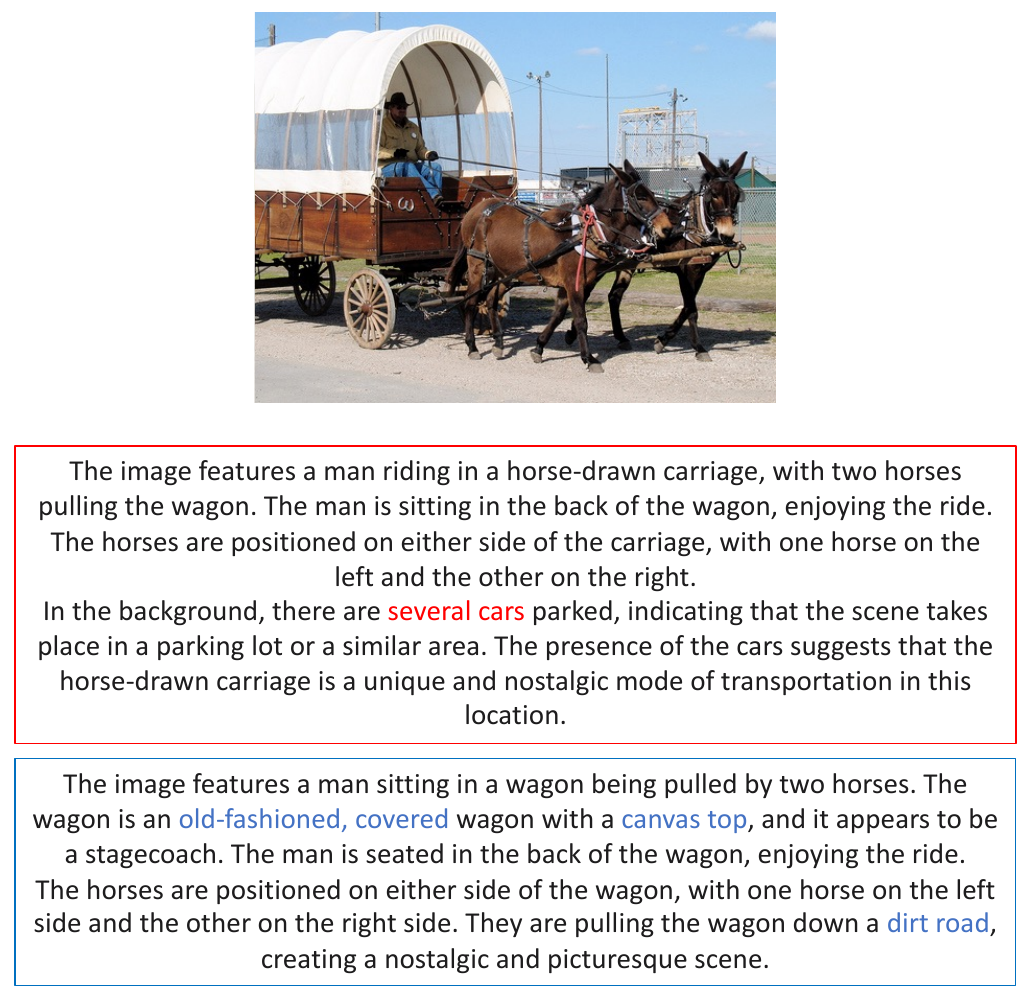}
\caption{Captions generated by LLaVA before and after ESREAL.}
\label{fig:9_qualitative_3}
\end{figure*}

%% file: Figures/10_gpt4v_prompt.tex
\begin{figure*}[!t]
\centering
\includegraphics[width=0.9\textwidth]{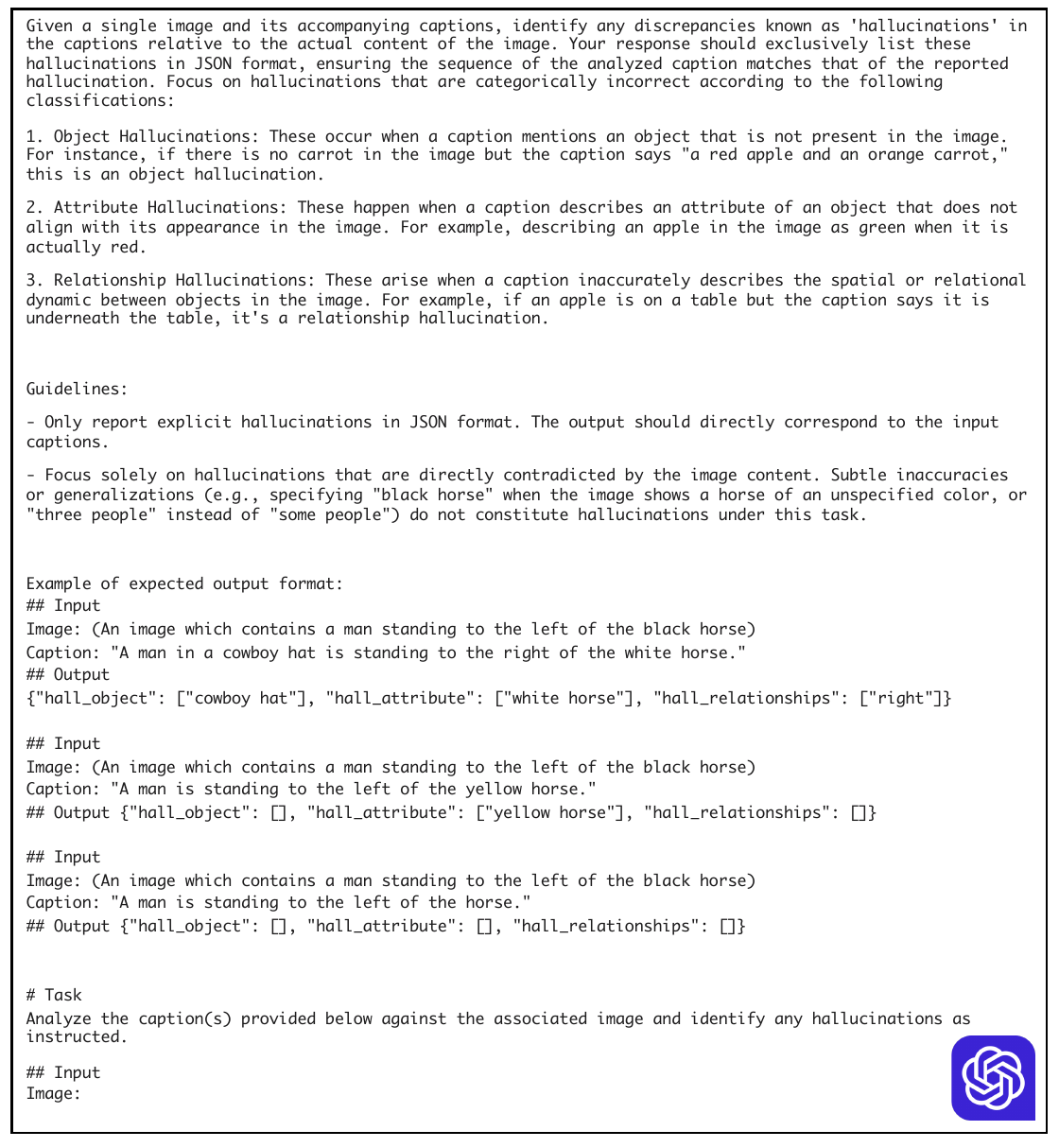}
\caption{Prompt for GPT-4V-aided evaluation.}
\label{fig:10_gpt4v_prompt}
\end{figure*}

%% file: Figures/11_gpt4v_output.tex
\begin{figure*}[!t]
\centering
\includegraphics[width=\textwidth]{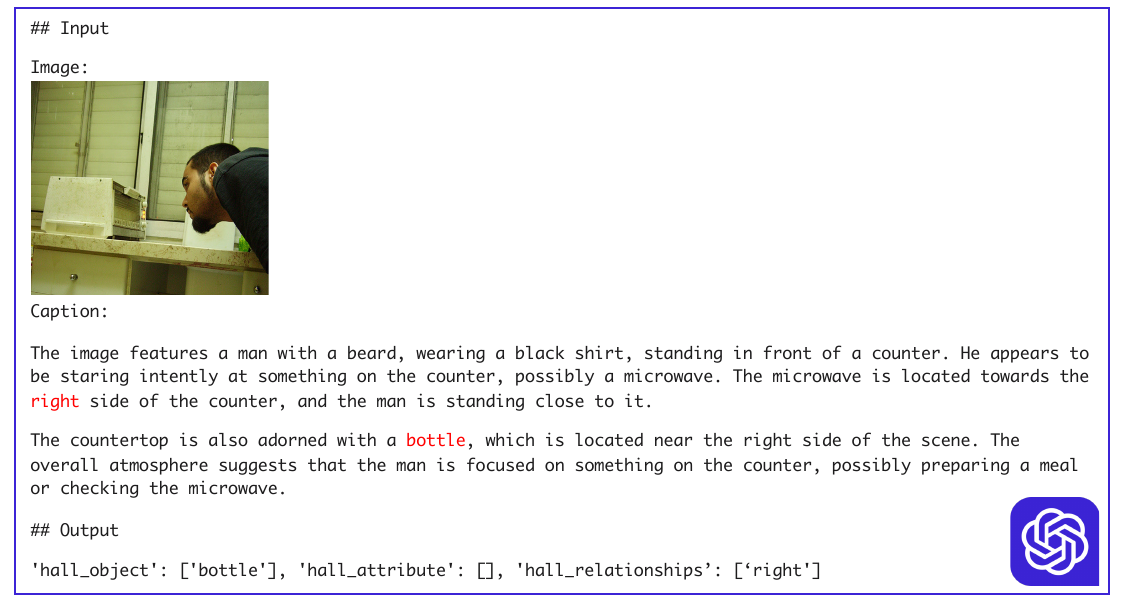}
\caption{Example output of GPT-4V-aided evaluation.}
\label{fig:11_gpt4v_output}
\end{figure*}

%% file: Figures/11_winrate_object.tex
\begin{figure*}[!t]
\centering
\includegraphics[width=\textwidth]{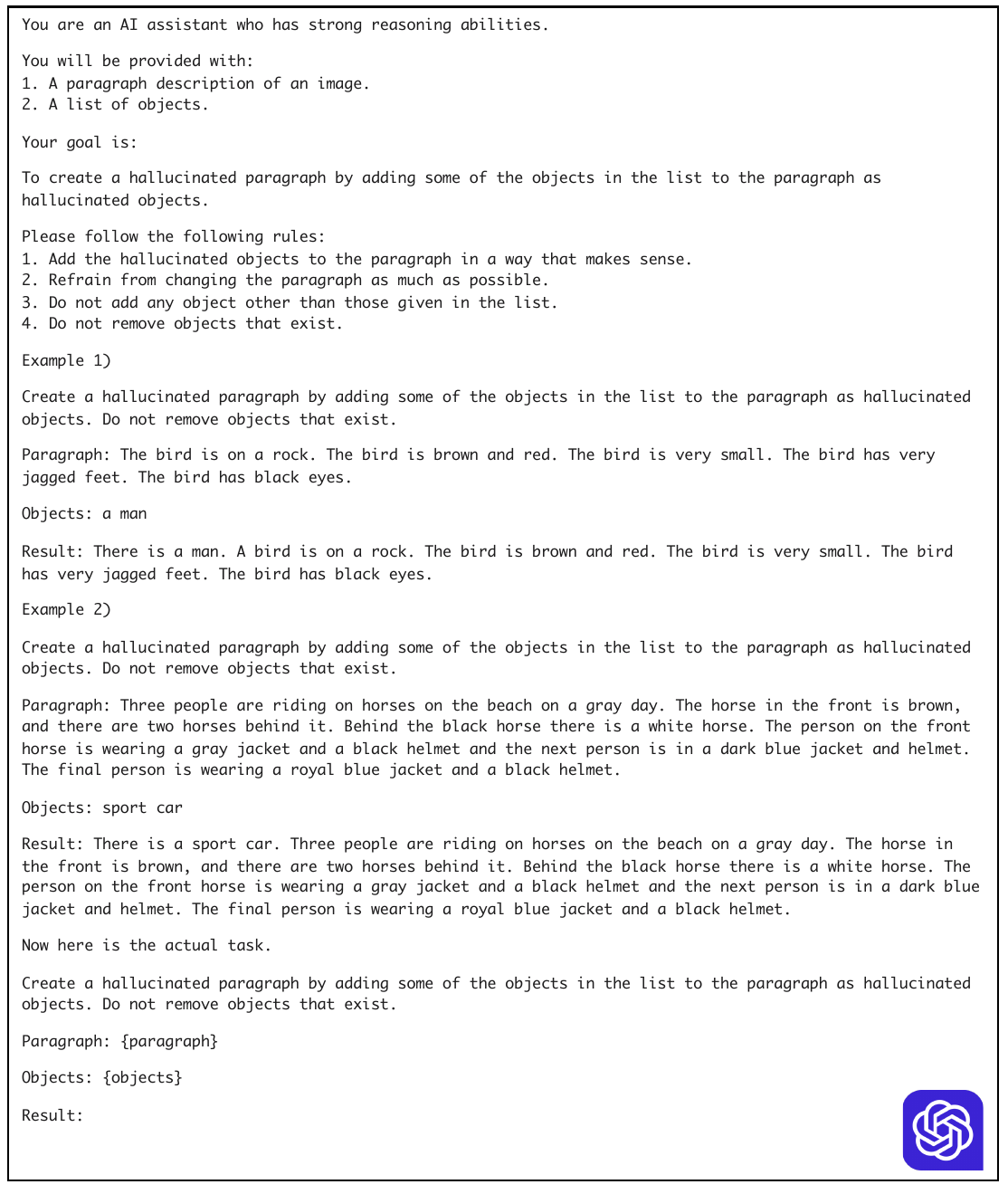}
\caption{Prompt for generating hallucinated captions with non-existent objects.}
\label{fig:11_winrate_object}
\end{figure*}

%% file: Figures/12_winrate_attribute.tex
\begin{figure*}[!t]
\centering
\includegraphics[width=\textwidth]{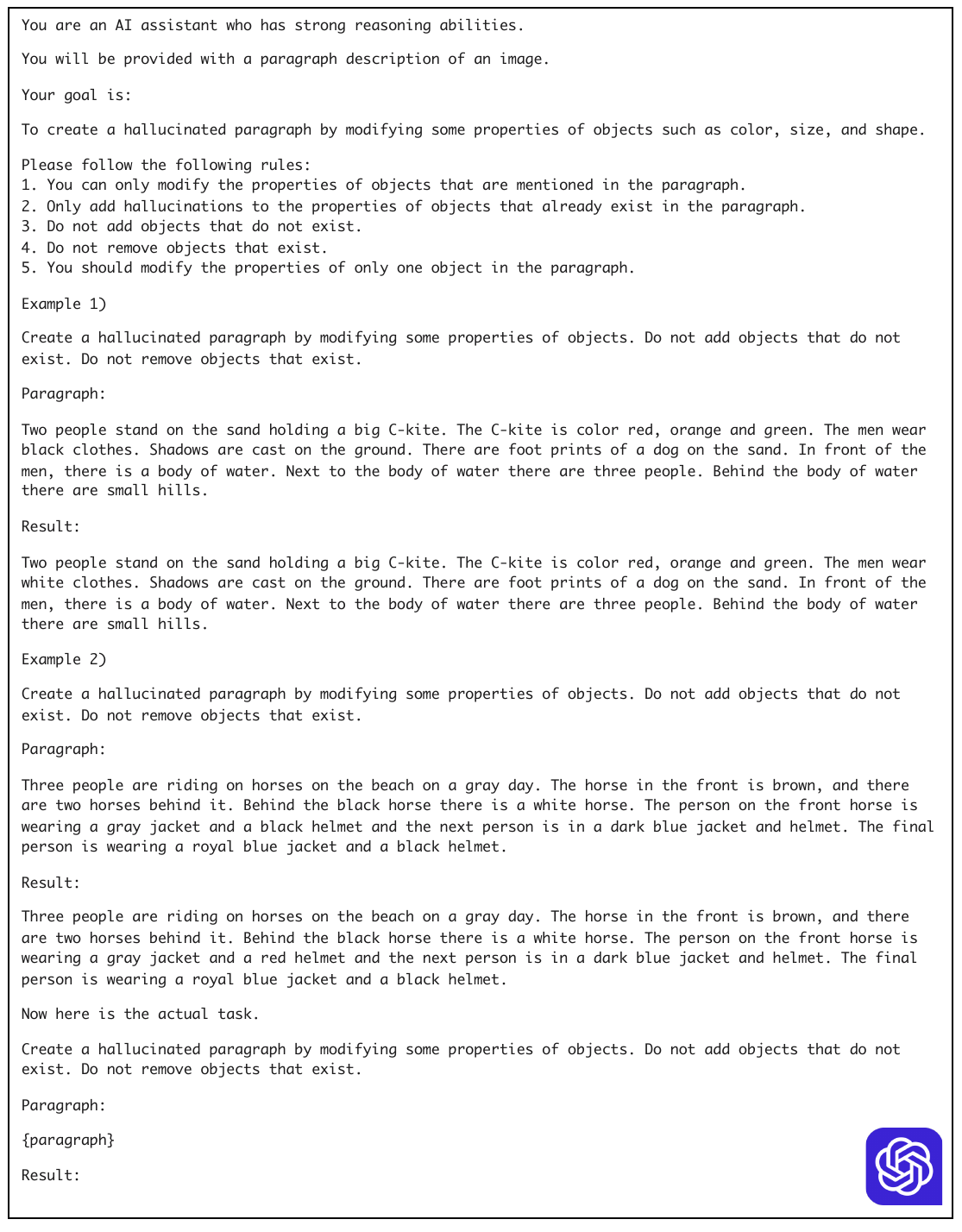}
\caption{Prompt for generating hallucinated captions with unfaithful object attributes.}
\label{fig:12_winrate_attribute}
\end{figure*}

%% file: Figures/13_winrate_relation.tex
\begin{figure*}[!t]
\centering
\includegraphics[width=\textwidth]{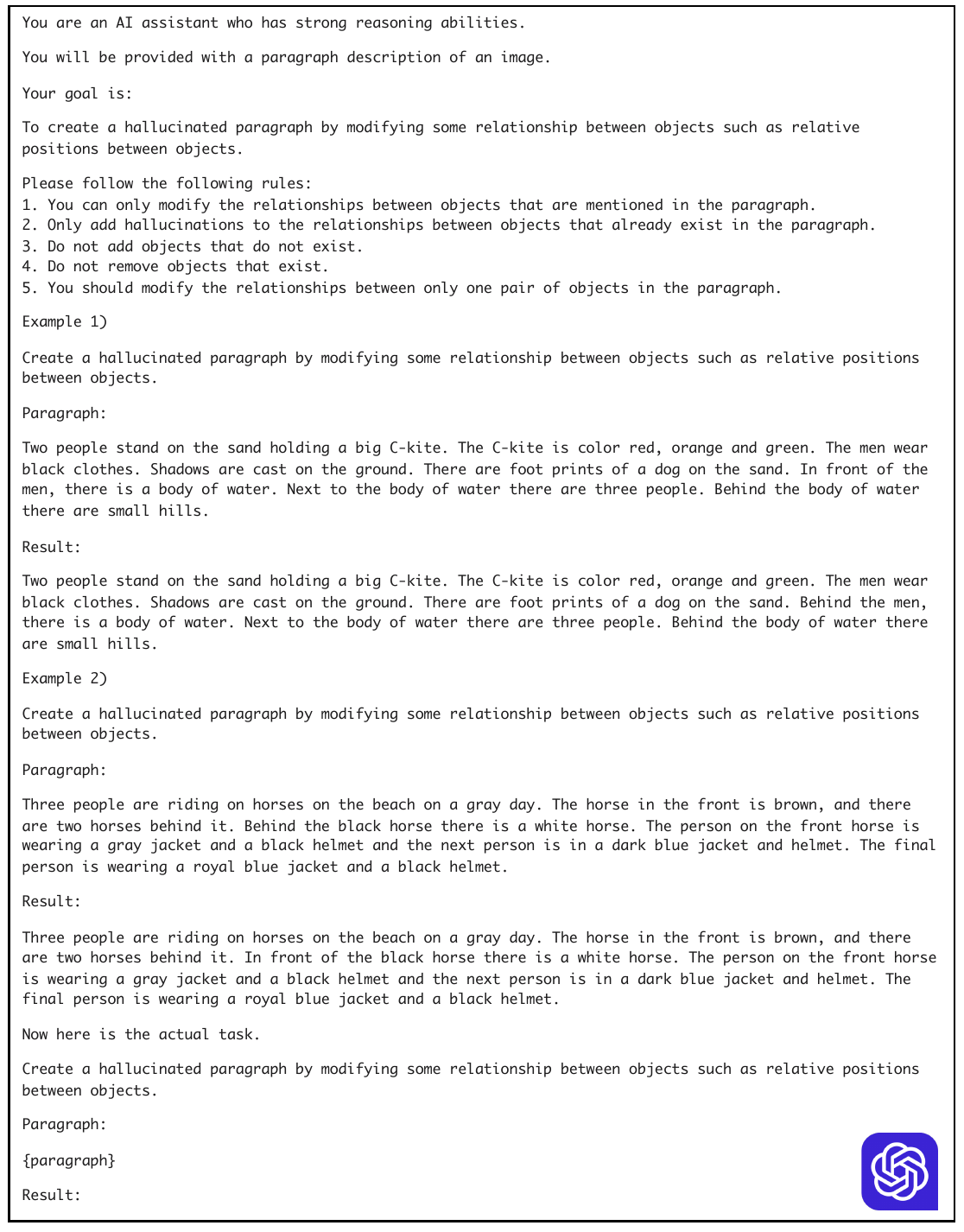}
\caption{Prompt for generating hallucinated captions with inaccurate relationships.}
\label{fig:13_winrate_relations}
\end{figure*}

%% file: Figures/14_winrate_output.tex
\begin{figure*}[!t]
\centering
\includegraphics[width=\textwidth]{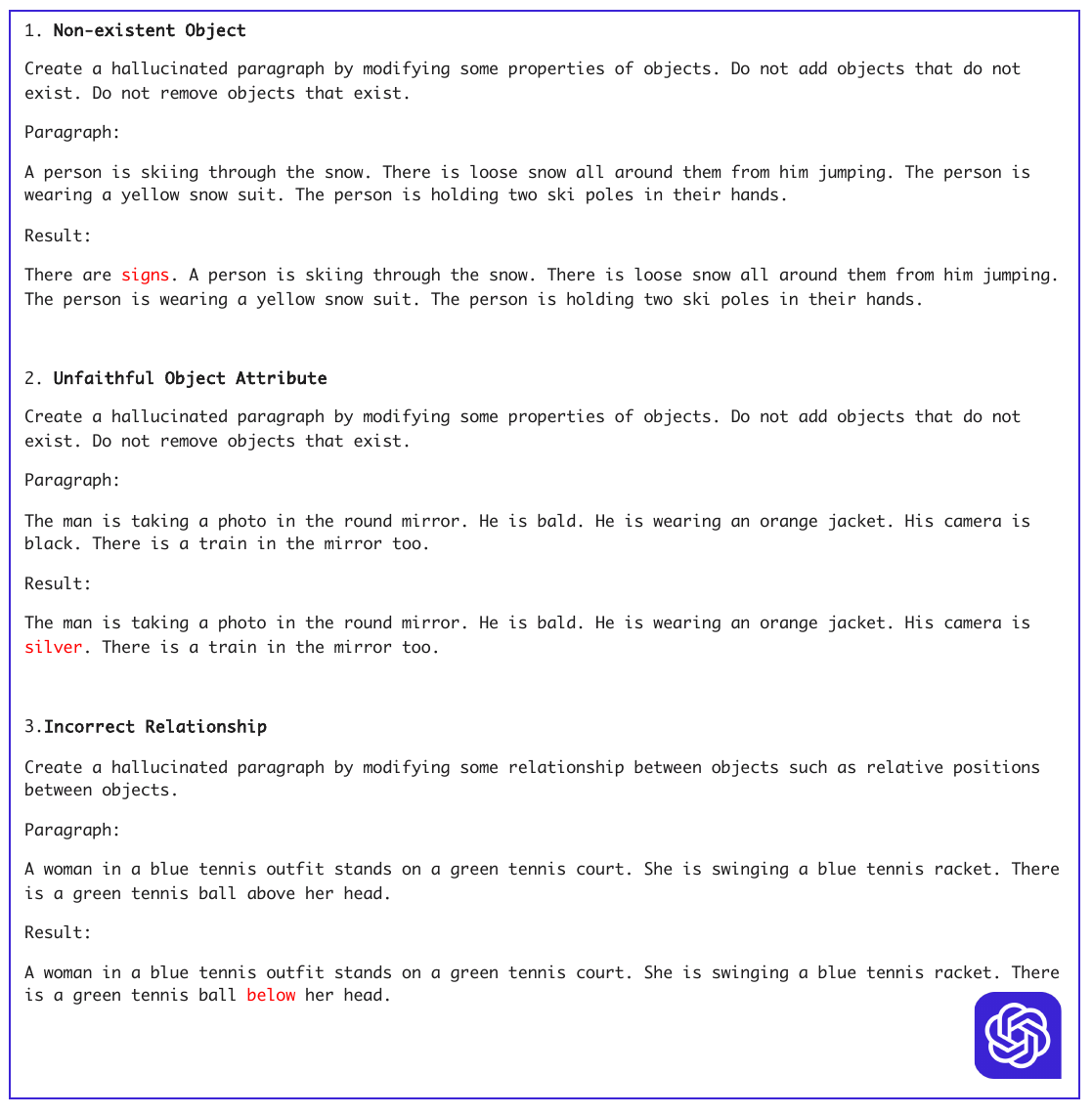}
\caption{Example output of generating hallucinated captions for Win Rate analysis.}
\label{fig:14_winrate_output}
\end{figure*}